%% file: main.tex
\PassOptionsToPackage{table}{xcolor}
\documentclass{article}

\usepackage{iclr2027_conference,times}

\input{math_commands.tex}

\input{preamble.tex}

\title{Strong yet Controllable LLM Abstention without Learning to Abstain}

\iclrfinalcopy

\author{Junwei Zhang\textsuperscript{\shieldicon\unvicon *} \hspace{0.7em}
Yifei Wang\textsuperscript{\capicon *} \hspace{0.7em}
Wanli Yang\textsuperscript{\shieldicon\unvicon} \hspace{0.7em}
Hexiang Tan\textsuperscript{\shieldicon\unvicon}  \hspace{0.7em}
\textbf{Rui Tang}\textsuperscript{\unvicon} \\ 
\textbf{Xin Chen}  \hspace{0.7em}
\textbf{Jingang Wang} \hspace{0.7em}
\textbf{Du Su}\textsuperscript{\shieldicon} \hspace{0.7em}
\textbf{Fei Sun}\textsuperscript{\shieldicon\,\tiny{\faIcon[regular]{envelope}}}\\
\textsuperscript{\shieldicon}State Key Laboratory of AI Safety, Institute of Computing Technology, CAS\\
\textsuperscript{\unvicon}University of Chinese Academy of Sciences\,\,\,  \textsuperscript{\capicon}Beijing Institute of Technology \\
\tt{zhangjunwei233@mails.ucas.ac.cn \,\,\, \textsuperscript{\tiny{\faIcon[regular]{envelope}}}sunfei@ict.ac.cn}
}

\begin{document}

\maketitle
\lhead{Preprint}

\begingroup
\begin{NoHyper}
\renewcommand*{\thefootnote}{*} 
\footnotetext{Junwei Zhang and Yifei Wang contributed equally to this work.}
\renewcommand*{\thefootnote}{\tiny\faIcon[regular]{envelope}}
\footnotetext{Corresponding author: Fei Sun (\href{mailto:sunfei@ict.ac.cn}{sunfei@ict.ac.cn})}
\end{NoHyper}
\endgroup

\renewcommand*{\thefootnote}{\arabic{footnote}}

\input{sections/00_abstract}
\input{sections/01_intro}
\input{sections/06_related_work}

\input{sections/02_preliminaries}
\input{sections/03_motivation}
\input{sections/04_methods}
\input{sections/05_experiments}

\input{sections/07_conclusion}
\clearpage
\input{sections/statements}

\bibliography{references}
\bibliographystyle{iclr2027_conference}

\appendix
\input{sections/appendix}

\end{document}

%% file: math_commands.tex
\usepackage{amsmath,amsfonts,bm}

\def\eqref#1{equation~\ref{#1}}

\def\1{\bm{1}}

\DeclareMathAlphabet{\mathsfit}{\encodingdefault}{\sfdefault}{m}{sl}
\SetMathAlphabet{\mathsfit}{bold}{\encodingdefault}{\sfdefault}{bx}{n}



%% file: preamble.tex
\usepackage{microtype}    
\usepackage[T1]{fontenc}  
\usepackage{xspace}       
\usepackage{xcolor}       
\usepackage{fontawesome5} 

\usepackage{mathtools}    
\usepackage{amssymb}      
\usepackage{bm}           

\usepackage{booktabs}     
\usepackage{multirow}     
\usepackage{tabularx}     
\usepackage{adjustbox}    
\usepackage{makecell}     

\usepackage{graphicx}     
\usepackage{subcaption}   
\usepackage{wrapfig}      

\usepackage[inline]{enumitem} 
\usepackage{pifont}           

\usepackage{listings}      
\usepackage{tcolorbox}     
\tcbuselibrary{breakable}  

\usepackage{tikz}
\usetikzlibrary{arrows.meta,shapes.geometric}

\usepackage{hyperref}      
\usepackage[capitalize]{cleveref}      

\newcommand{\ie}{\textit{i.e.}\xspace}
\newcommand{\eg}{\textit{e.g.}\xspace}

\newcommand{\unvicon}{\scalebox{0.8}{\tiny\faIcon{university}}}
\newcommand{\shieldicon}{\scalebox{0.8}{\tiny\faIcon{shield-alt}}}

\newcommand{\capicon}{\scalebox{0.8}{\tiny\faIcon{graduation-cap}}}

%% file: sections/00_abstract.tex
\begin{abstract}

Large language models (LLMs) can reduce hallucination by abstaining from questions they would likely get wrong.
Abstention can be implemented outside the model (\textit{selective answering}) using a confidence threshold that controls the answer rate (\textit{coverage}), or learned inside it (\textit{abstention training}), which can perform better but fixes coverage once trained.
To compare the two, we match their coverage and decompose \textit{risk} (error rate) into \textit{answering ability} and \textit{distinguishing ability}, \ie, how well the model answers and how accurately it abstains, respectively.
The decomposition shows that abstention training performs better mainly by answering better: with strong confidence estimators, the two have nearly equal distinguishing ability.
This motivates \textbf{T}rain for \textbf{A}ccuracy, then \textbf{P}ost-hoc \textbf{S}elect (\textbf{TAPS}), a modular approach that keeps coverage controllable, needs only standard post-training and an off-the-shelf confidence estimator, and benefits from advances in either component.
Across three LLMs and three factual QA benchmarks, TAPS with a probing-based confidence estimator attains 1.5--9.4 percentage points lower mean risk than three representative abstention training methods at matched coverage.
Within TAPS, accuracy training often weakens the model's own confidence signals, yet refitting the estimator largely preserves distinguishing ability.
Together, these results establish TAPS as a strong yet controllable baseline and raise the bar for future abstention training: sacrificing control over coverage should yield gains beyond what can be achieved by optimizing answering and abstaining separately.

\end{abstract}

%% file: sections/01_intro.tex
\section{Introduction}
\label{sec:introduction}

Hallucination remains a central challenge to the reliable deployment of large language models (LLMs) \citep{ziweiji2023survey}.
A direct way to mitigate this problem is \emph{abstention}: declining questions the model is likely to answer incorrectly \citep{wen2025know-your-limits}.
Abstention trades quantity for quality: the model answers fewer questions, but the answers it gives are more likely to be correct.
Existing methods fall into two paradigms according to where the abstention decision is made.
\textit{Selective answering} decides outside the model, after generation: the model first generates an answer, and a separate confidence estimator withholds it if its confidence falls below a threshold \citep{kamath-etal-2020-selective, xin-etal-2021-art, yoshikawa-okazaki-2023-selective}.
\textit{Abstention training} instead internalizes the decision: during generation, the model itself learns both how to answer correctly and when to abstain \citep{zhang-etal-2024-r, wei2025truthrl, zhou2026whensilence, zeng2026ucpo}.


The two paradigms exhibit complementary strengths and limitations. 
Selective answering is simple and flexible.
It reduces abstention to confidence estimation without modifying the model, and adjusting the threshold controls how conservatively it answers, thereby allowing a single model to serve applications with different tolerances for error \citep{liu2022incorporating, srinivasan-etal-2024-selective}.
In contrast, abstention training has shown promising results in reducing hallucination \citep{wei2025truthrl}, though each trained model is tied to a single, intrinsic answer rate.
Moreover, recent studies find that even under different training incentives, the answer rates of such models tend to collapse into a narrow, conservative range \citep{che2026abstention, zeng2026ucpo}, as we also observe (\S\ref{sec:rq1}, Appendix~\ref{app:coverage-control}).
\citet{che2026abstention} further attribute this collapse to the optimization dynamics of abstention training itself. 
This lack of control also persists at inference time, where prompting a trained model to vary its answer rate remains challenging \citep{wu2026behavioral}.
A natural question, then, is \textit{whether abstention can be both strong and controllable}.

\begin{wrapfigure}[14]{r}{0.35\textwidth}
    \vspace{-12pt}
    \centering
    \includegraphics[width=\linewidth]{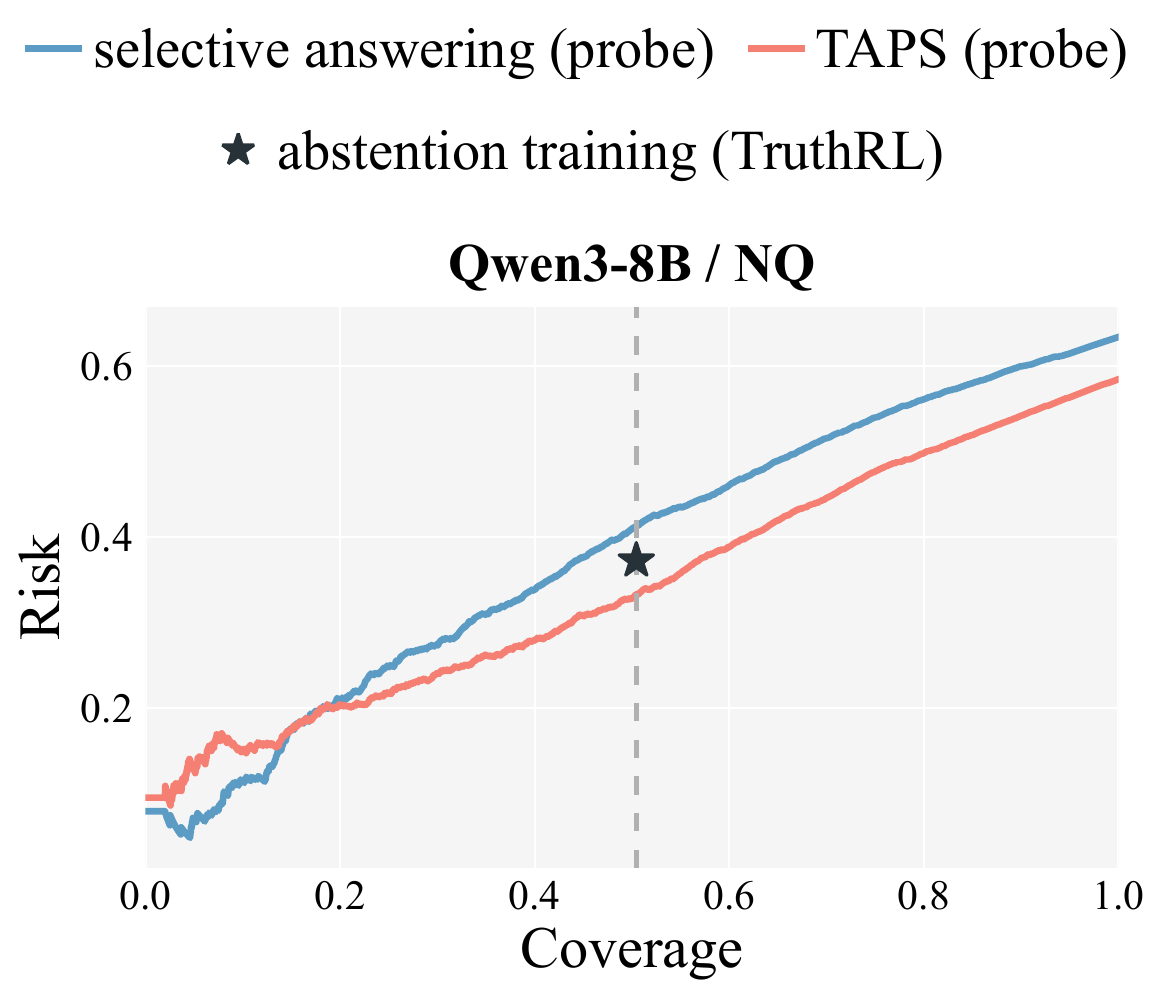}
    \captionsetup{skip=3pt}
    \caption{Two different abstention paradigms and TAPS (\textbf{ours}).
    }
    \label{fig:risk-coverage-demo}
\end{wrapfigure}

Answering this question first requires comparing the two paradigms on common ground, since they have developed largely in parallel and are evaluated differently.
Abstention training studies typically report a single score, such as accuracy, hallucination rate, their difference, or accuracy on answered questions alone \citep{zhang-etal-2024-r, wei2025truthrl, zeng2026ucpo}.
Consider one model that answers 5\% of questions, 99\% of them correctly, and another that answers 80\%, 90\% of them correctly.
The first has the lower hallucination rate (0.05\% vs.\ 8\%) and the second the higher accuracy (72\% vs.\ 4.95\%).
Yet neither score reveals which abstains better, since the two models operate at different answer rates: the second might also achieve 99\% accuracy if limited to its 5\% most confident answers.
Thus, comparing abstention quality requires holding the answer rate fixed.
We therefore adopt the evaluation protocol of selective prediction \citep{geifman2017selective}, which separately tracks \textit{coverage}, the fraction of questions answered, and \textit{risk}, the error rate among them.
In this risk--coverage plane, selective answering traces a curve as its threshold varies, whereas an abstention-trained model occupies a single point (\Cref{fig:risk-coverage-demo}), so we compare their risks at that point's coverage (the vertical line), which we call \textit{matched coverage}.



Our preliminary experiments confirm that TruthRL \citep{wei2025truthrl}, a representative RL-based abstention training method, attains lower mean risk at matched coverage than selective answering applied to the same starting model (\eg, the star point below the blue line in \Cref{fig:risk-coverage-demo}).
We note that, at fixed coverage, lower risk can come from only two sources: generating fewer wrong answers or withholding answers that would be wrong rather than correct ones.
This motivates an exact decomposition of risk into these two sources, which we term \emph{answering ability} and \emph{distinguishing ability}, respectively.
The decomposition shows that TruthRL's advantage in distinguishing ability shrinks as selective answering uses stronger confidence estimators and even vanishes on average with the strongest.
What remains comes entirely from answering ability.
Despite its name, then, what sets abstention training apart from selective answering is how it answers, not how it abstains.

This raises a question: \textit{what if selective answering were built on an equally strong answerer}? 
We therefore introduce \textit{\textbf{T}rain for \textbf{A}ccuracy, then \textbf{P}ost-hoc \textbf{S}elect} (\textbf{TAPS}), a simple two-stage approach that first trains the model solely for answer accuracy and then leaves the abstention decision to a post-hoc confidence estimator (\eg, red curve in \Cref{fig:risk-coverage-demo}).
Because the decision stays outside the model, TAPS preserves controllability: coverage is adjusted through a threshold, just as in selective answering.
It is also practical and modular: task-specific post-training is already standard practice, and abstention then requires only an off-the-shelf confidence estimator rather than a specialized training objective.
Moreover, because the two stages draw on well-developed and largely independent fields, TAPS can directly incorporate advances in either.

We evaluate TAPS on three representative open-source LLMs and three factual QA benchmarks against three abstention training methods, R-Tuning \citep{zhang-etal-2024-r}, TruthRL \citep{wei2025truthrl}, and UCPO \citep{zeng2026ucpo}, all trained from the same starting model on the same data as TAPS. 
At matched coverage, TAPS with a lightweight probing-based estimator \citep{orgrad-2025-llmsknow} attains 1.5 percentage points (pp) lower mean risk than TruthRL (winning 17 of 27 comparisons) and 8.1 and 9.4 pp lower than R-Tuning and UCPO, respectively. 
Crucially, it does so without sacrificing controllability, whereas TruthRL's coverage barely moves even when it is retrained with different abstention rewards.
Further analysis highlights the complementary roles of TAPS's two stages.
At the first stage, accuracy training emerges as the main contributor, lowering risk in 81 of 90 matched-coverage comparisons relative to selective answering without it, even when it weakens the model's own confidence signals.
At the second stage, a supervised probing-based estimator refit on the trained model largely preserves distinguishing ability despite this weakening, and more generally, stronger estimators yield lower risk.
These results do not argue against abstention training; rather, they establish TAPS as a strong baseline for isolating the gains specific to learning to abstain.

Our main contributions are summarized as follows:
\begin{itemize}[leftmargin=25pt, topsep=1pt, itemsep=0pt]
    \item We compare abstention training and selective answering under a common risk--coverage evaluation protocol with controlled training conditions, and derive an exact decomposition of their risk into answering ability and distinguishing ability.
    \item This decomposition shows abstention training learns when to abstain and how to answer better, but only the latter sets it apart from selective answering with a strong confidence estimator.
    \item We establish TAPS as a strong and controllable baseline. Without any abstention-specific objective, it attains lower mean risk than three representative abstention training methods at matched coverage while keeping coverage adjustable.
\end{itemize}

%% file: sections/06_related_work.tex
\section{Related Work}
\label{sec:related}

LLM abstention inherits a distinction between internalizing and externalizing the reject decision from classical selective prediction, which we review in Appendix~\ref{app:selective-prediction}.

\paragraph{Selecting answers post-hoc to abstain.}
Selective answering externalizes abstention by thresholding answer confidence \citep{kamath-etal-2020-selective,xin-etal-2021-art,yoshikawa-okazaki-2023-selective}, which allows flexible control of the answer rate.
Work in this field mainly focuses on improving confidence estimation.
One line of research estimates confidence from a fixed LLM and can be divided into two groups.
\emph{Unsupervised methods} either prompt the model to self-evaluate, as in P(True) \citep{kadavath2022language} and verbalized confidence \citep{tian2023calibration}, or sample multiple responses and measure answer agreement \citep{cole-etal-2023-selectively} or semantic entropy \citep{semantic_entropy,farquhar2024detecting}.
\emph{Supervised methods} fit external predictors on handcrafted features such as text length and prediction probabilities \citep{kamath-etal-2020-selective,varshney-etal-2022-improving}, or train correctness probes on hidden states \citep{orgrad-2025-llmsknow}.
Another line trains the LLM itself to express calibrated confidence \citep{lin2022teaching,xu-etal-2024-sayself,damani2026beyond-binary-reward} or directly optimizes selective performance \citep{luo2026aligning}.
Closest to TAPS, ASPIRE \citep{chen-etal-2023-adaptation} adapts the answerer via soft prompt tuning and learns a self-evaluation score on its own sampled answers.
However, ASPIRE aims to improve selective prediction itself, whereas TAPS revisits abstention training and decomposes where its gains come from.
Moreover, TAPS is a general framework compatible with standard post-training and any confidence estimator, whereas ASPIRE is tied to an SFT-style training paradigm and a self-evaluation score, and was developed only for small, non-instruction-tuned LMs (up to 2.7B).

\paragraph{Training language models to abstain.}
In contrast, abstention training internalizes the abstention decision into the model policy.
Early \textit{supervised fine-tuning methods} first construct a dataset with ``I don't know'' labels: they probe the model on training questions and mark those beyond its knowledge boundary as abstentions, \eg, R-Tuning \citep{zhang-etal-2024-r}, honesty-oriented SFT \citep{yang2024seal}, and Idk-SFT \citep{cheng2024canai}.
This pre-sampling step is costly and ties the labels to the model checkpoint at which they were collected.
More recent \textit{reinforcement-learning methods} avoid it by scoring abstention directly on the model's own on-policy rollouts with outcome rewards \citep{wei2025truthrl,mohamadi2025honesty,zhou2026whensilence}, and achieve stronger results in reducing hallucination.
For instance, TruthRL assigns $+1$, $0$, and $-1$ to correct answers, abstentions, and errors, respectively.
However, the resulting policies tend to collapse into narrow answer-rate ranges \citep{che2026abstention,zeng2026ucpo}.
UCPO \citep{zeng2026ucpo} and TIAR \citep{pan2026tiar} attempt to mitigate this by replacing static abstention advantages with dynamic ones, yet \citet{che2026abstention} trace the collapse to vanishing reward and KL gradients, suggesting that such fixes cannot fully resolve it.


%% file: sections/02_preliminaries.tex
\section{Abstention as Selective Prediction}
\label{sec:preliminaries}

Abstention training pursues the same goal as selective answering at its core, but the metrics it adopts cannot support a fair comparison between the two paradigms.
In this section, we first explain this limitation and then formalize both paradigms under the standard selective prediction framework, whose risk--coverage evaluation protocol enables fair comparison via risk at matched coverage.



Abstention mitigates hallucination by withholding answers that are likely to be wrong.
Formally, consider a dataset  $\mathcal{D}=\{(x_i,y_i^*)\}_{i=1}^{N}$ of $N$ questions, where \(x_i\) is a question and \(y_i^*\) is its reference answer.
An abstention method takes a question $x_{i}$ and either returns an answer $\hat{y}_{i}$ or abstains, partitioning the dataset into an answered set $\mathcal{A}$ and an abstained set $\mathcal{R}$.
Ideally, all correct answers should remain in \(\mathcal A\) and all others should fall into \(\mathcal R\).
Real methods fall short in both directions,
causing a trade-off between the fraction of questions answered and the error rate among those answers.

Metrics commonly adopted by prior abstention training studies fail to fully characterize this trade-off.
Abstention is neither correct nor incorrect, but any single score over all questions must assign it a value relative to the other two.
Prior studies do so implicitly \citep{yang2024seal, cheng2024canai}: overall accuracy values an abstention as an incorrect answer, hallucination rate values it as a correct answer, and composites such as accuracy minus hallucination rate \citep{wei2025truthrl} value it midway between the two.
Under any such valuation, a difference in score need not reflect better abstention decisions.
It may simply reflect a different amount of abstention: withholding answers at random, for instance, lowers the hallucination rate without improving a single decision.
Accuracy over the answered questions alone \citep{zhang-etal-2024-r, zeng2026ucpo} sidesteps the valuation by ignoring abstentions, but then omits the coverage they cost.
These metrics thus suffice to describe one policy at its native coverage, but cannot fairly compare policies with different coverages.

To fully characterize this trade-off and enable fair comparison, we keep its two sides separate, as \textit{coverage}, \ie, how many questions are answered, and \textit{risk}, \ie, how often those answers are wrong:
\begin{equation}
c=\frac{|\mathcal{A}|}{N},\quad r=\frac{1}{|\mathcal{A}|}\sum_{i\in\mathcal{A}} e_i,
\end{equation}
where $e_i=1$ if $\hat{y}_i$ is incorrect relative to $y_i^*$, and $0$ otherwise.
This is the standard formulation of selective prediction \citep{geifman2017selective}, in which a predictor is equipped with a reject option and evaluated by these two metrics jointly.
LLM abstention can be viewed as selective prediction under a generative setup, with the answering model as the predictor and the abstention decision as the reject option.
Within this space, different abstention behaviors induce different operating points \((c,r)\).
Let $r(c)$ denote the risk at coverage \(c\). A better method attains lower $r(c)$ across coverage levels, which is summarized by lower \textbf{area under the risk--coverage curve (AURC)}:
\begin{equation}
\mathrm{AURC} = \int_{0}^{1} r(c)\,dc.
\label{eq:aurc}
\end{equation}
Since any abstention method, however implemented, reduces to points in this space, the formulation provides common ground for comparing the two paradigms.
For selective answering methods, since sweeping the confidence threshold traces a full risk--coverage curve, they can be directly compared using AURC.
An abstention-trained policy, in contrast, is fixed once trained and typically yields only a single native operating point $(c_{\mathrm{train}},r_{\mathrm{train}})$.
Therefore, instead of estimating AURC, we compare the two paradigms \textit{at matched coverage}: we adjust the acceptance threshold of selective answering to match $c_{\mathrm{train}}$, and compare the resulting risk with $r_{\mathrm{train}}$, which we denote as \textbf{Risk@Cov}.




%% file: sections/03_motivation.tex
\section{Understanding the Advantage of Abstention Training}
\label{sec:motivation}
With the two abstention paradigms placed on common ground, we ask whether abstention training outperforms selective answering, and if so, where this advantage comes from.

\subsection{Behavioral Decomposition of Abstention Performance}
\label{sec:preliminary-comparison}

\begin{figure*}[t]
    \centering
    \input{figures/sankey_comparison}
    \caption{
    \textbf{Behavioral decomposition of abstention performance at matched coverage $c$.}
    \subref*{fig:sankey-analysis-decomposition} Decomposing risk into answering and distinguishing abilities.
    \subref*{fig:sankey-analysis-comparison} On Qwen2.5-7B-Instruct/TriviaQA at matched coverage $c=72.7\%$,
    stronger selective answering (S.E.) closes the distinguishing ability gap ($G_{\mathrm{dis}}$) with TruthRL, but TruthRL's edge in answering ability ($G_{\mathrm{ans}}$) still yields lower risk.
    }
    \label{fig:sankey-analysis}
\end{figure*}
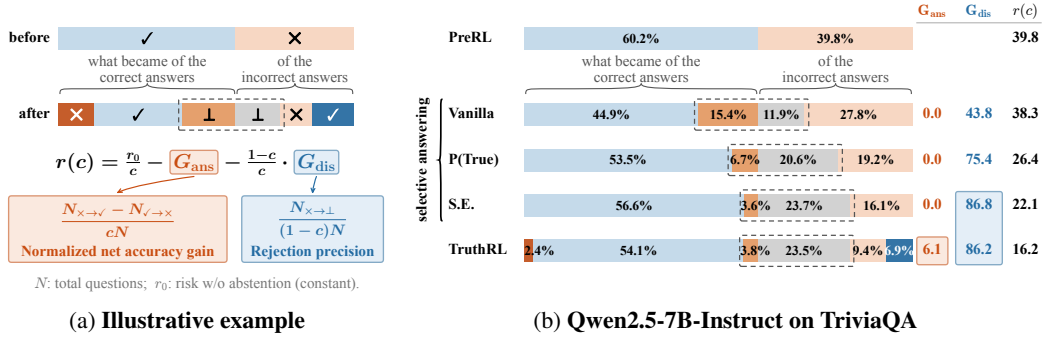

We first build a theoretical lens for advantage analysis.
Following the Risk@Cov comparison protocol established in \Cref{sec:preliminaries}, we analyze what determines risk at a fixed coverage through the behavioral transitions illustrated in \Cref{fig:sankey-analysis-decomposition}.
Before applying an abstention method, each question is answered either correctly ($\checkmark$) or incorrectly ($\times$).
After applying it, each answer flows into one of three outcomes: correct, incorrect, or abstained ($\perp$).
Let $N_{a\to a'}$ denote the number of questions transitioning from $a\in\{\checkmark,\times\}$ to $a'\in\{\checkmark,\times,\perp\}$.
These transitions reduce the number of remaining errors in two ways.
First, answer generation may change: correcting an initially incorrect answer ($N_{\times\to\checkmark}$) removes an error, while breaking an initially correct answer ($N_{\checkmark\to\times}$) adds one, for a net effect of $N_{\times\to\checkmark}-N_{\checkmark\to\times}$.
Second, abstention removes an error whenever an initially incorrect answer is withheld, contributing $N_{\times\to\perp}$.
Starting from $N_{\times(\mathrm{before})}$ incorrect answers, the final number of errors is thus
\begin{equation}
N_{\times(\mathrm{after})}
=
N_{\times(\mathrm{before})}
-
\left(
N_{\times\to\checkmark}
-
N_{\checkmark\to\times}
\right)
-
N_{\times\to\perp}.
\label{eq:error-decomposition}
\end{equation}

At coverage $c$, exactly $cN$ questions remain answered and $(1-c)N$ are abstained. Dividing \Cref{eq:error-decomposition} by $cN$ gives the risk under coverage $c$:
\begin{equation}
r(c)
=
\frac{1}{c}
\cdot
\underbrace{\frac{N_{\times(\mathrm{before})}}{N}}_{r_0}
-
\underbrace{
\frac{N_{\times\to\checkmark}-N_{\checkmark\to\times}}{cN}
}_{G_{\mathrm{ans}}}
-
\frac{1-c}{c}
\cdot
\underbrace{
\frac{N_{\times\to\perp}}{(1-c)N}
}_{G_{\mathrm{dis}}}.
\label{eq:risk-decomposition}
\end{equation}

Here, $r_0$ is the risk without abstention, $G_{\mathrm{ans}}$ is the normalized net accuracy gain (corrected minus broken in \Cref{fig:sankey-analysis}), and $G_{\mathrm{dis}}$ is the fraction of abstained questions that were initially incorrect, \ie, the rejection precision (the proportion of correct rejections in the abstained block in \Cref{fig:sankey-analysis}).
We refer to the latter two as \emph{answering ability} and \emph{distinguishing ability}, respectively.
When methods are applied to the same model and are compared at the same coverage, $r_0/c$ and $(1-c)/c$ are fixed, so \Cref{eq:risk-decomposition} gives an exact behavioral decomposition of Risk@Cov into $G_{\mathrm{ans}}$ and $G_{\mathrm{dis}}$.

\subsection{Preliminary Findings}

We next examine whether abstention training lowers risk relative to selective answering at matched coverage, and use the decomposition to trace where any advantage comes from.
We compare TruthRL~\citep{wei2025truthrl}, a representative RL-based abstention training method, against selective answering applied to the same starting model (\textit{PreRL}).
For selective answering, we consider Vanilla log-probability, P(True), and Semantic Entropy (S.E.) as confidence estimators, and adjust their thresholds to match TruthRL's native coverage.
TruthRL uses its default abstention reward $r_{\mathrm{abs}}=0$.
We perform this comparison across three models and three factual QA datasets.
The detailed setup is described in \Cref{sec:experimental-setup}.
\Cref{fig:sankey-analysis-comparison} illustrates the result of Qwen2.5-7B-Instruct on TriviaQA, while Appendix~\ref{app:extended-behavior-decomposition-results} reports the complete results.

\paragraph{Stronger confidence estimation closes the distinguishing gap.}
\Cref{fig:sankey-analysis-comparison} shows that, against selective answering with Vanilla confidence, TruthRL achieves both a positive $G_{\mathrm{ans}}$ and a higher $G_{\mathrm{dis}}$, so improvements in both abilities contribute to its lower risk.
However, this apparent distinguishing advantage shrinks as the confidence estimator grows stronger.
As reported in Appendix~\ref{app:extended-behavior-decomposition-results}, averaged across the nine model--dataset settings, $G_{\mathrm{dis}}$ rises from 59.47\% with Vanilla to 84.64\% with S.E., essentially matching TruthRL's 84.29\% average.
The distinguishing advantage of abstention training thus largely disappears once compared against a strong post-hoc selector.

\paragraph{The remaining gap lies in answering ability.}
As shown in \Cref{fig:sankey-analysis-comparison}, selective answering leaves the generated answers unchanged and therefore has $G_{\mathrm{ans}}=0$ by construction, whereas TruthRL directly benefits from it and attains lower risk.
As reported in Appendix~\ref{app:extended-behavior-decomposition-results}, TruthRL achieves a positive $G_{\mathrm{ans}}$ in every setting, averaging 5.07\%.
Together, these findings suggest that \textit{the performance advantage of abstention training over strong selective answering methods primarily reflects improved answering ability rather than substantially better distinguishing ability.}

%% file: figures/sankey_comparison.tex
\begingroup
\definecolor{sankeycorrect}{RGB}{197,218,234}
\definecolor{sankeyincorrect}{RGB}{247,215,195}
\definecolor{sankeyrejected}{RGB}{210,210,210}
\definecolor{sankeyfalse}{RGB}{229,160,109}
\definecolor{sankeycorrected}{RGB}{52,116,166}
\definecolor{sankeybroken}{RGB}{197,93,43}
\newcommand{\sankeykey}[2]{%
    \tikz[baseline=-0.5ex]{\fill[#1] (0,0) rectangle (0.20,0.14);}%
    ~#2}
\resizebox{\linewidth}{!}{%
    \scriptsize
    \sankeykey{sankeycorrect}{answered correctly}\quad
    \sankeykey{sankeyincorrect}{answered incorrectly}\quad
    \tikz[baseline=-0.5ex]{\draw[dashed,draw=black!65,line width=0.5pt]
        (0,0) rectangle (0.20,0.14);}~abstained\quad
    \sankeykey{sankeyrejected}{correct rejection}\quad
    \sankeykey{sankeyfalse}{false rejection}\quad
    \sankeykey{sankeycorrected}{corrected}\quad
    \sankeykey{sankeybroken}{broken}%
}\par\medskip
\captionsetup[subfigure]{labelformat=simple,hypcap=true,font=footnotesize}
\renewcommand{\thesubfigure}{(\alph{subfigure})}
\begin{subfigure}[t]{0.374\linewidth}
    \centering
    \includegraphics[width=\linewidth]{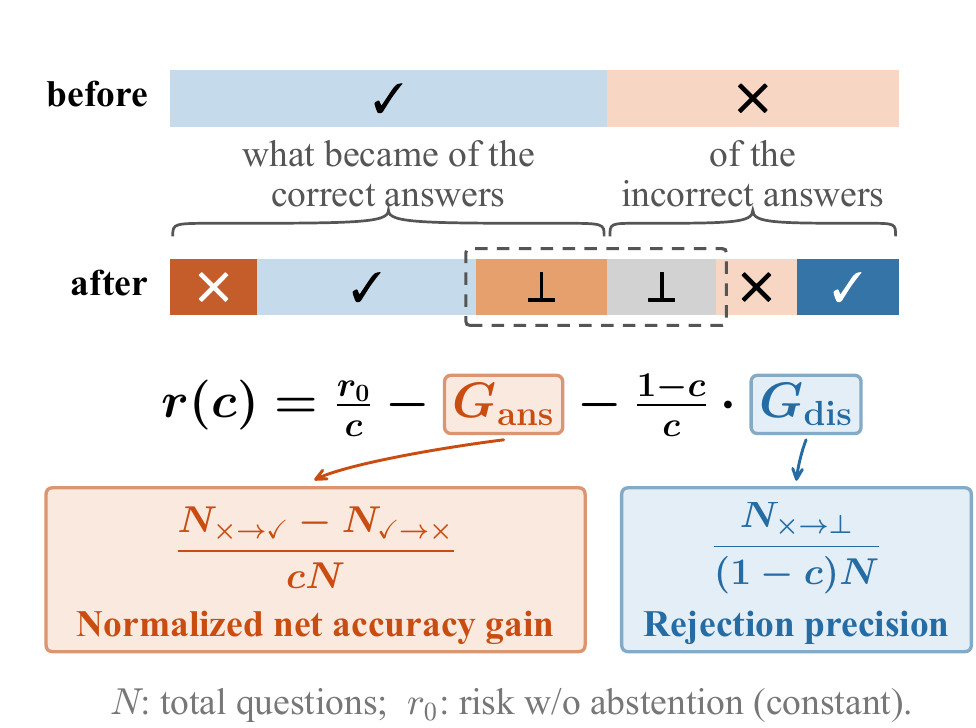}
    \caption{\textbf{Illustrative example}}
    \label{fig:sankey-analysis-decomposition}
\end{subfigure}%
\hfill
\begin{subfigure}[t]{0.606\linewidth}
    \centering
    \includegraphics[width=\linewidth]{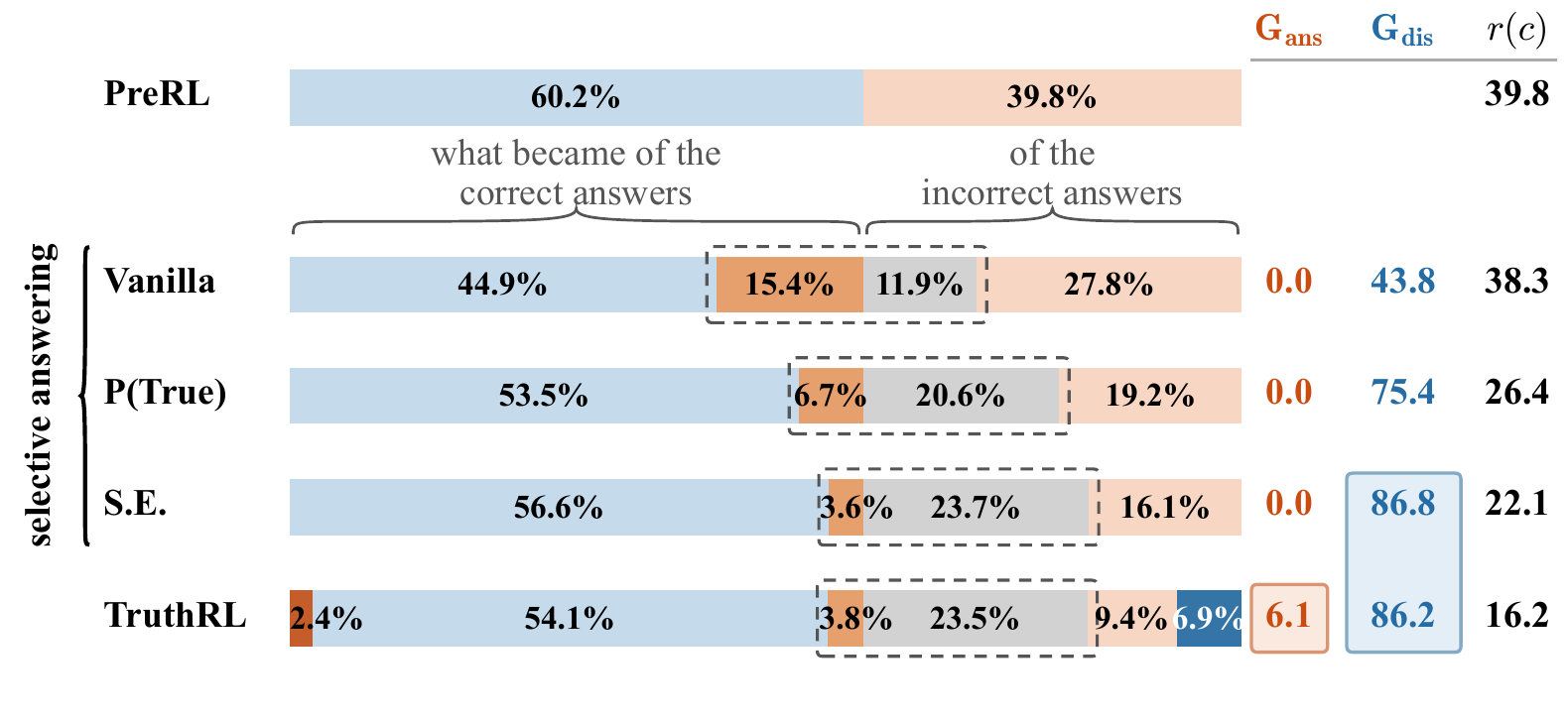}
    \caption{\textbf{Qwen2.5-7B-Instruct on TriviaQA}}
    \label{fig:sankey-analysis-comparison}
\end{subfigure}%
\endgroup

%% file: sections/04_methods.tex
\section{Train for Accuracy, then Post-hoc Select}
\label{sec:taps}

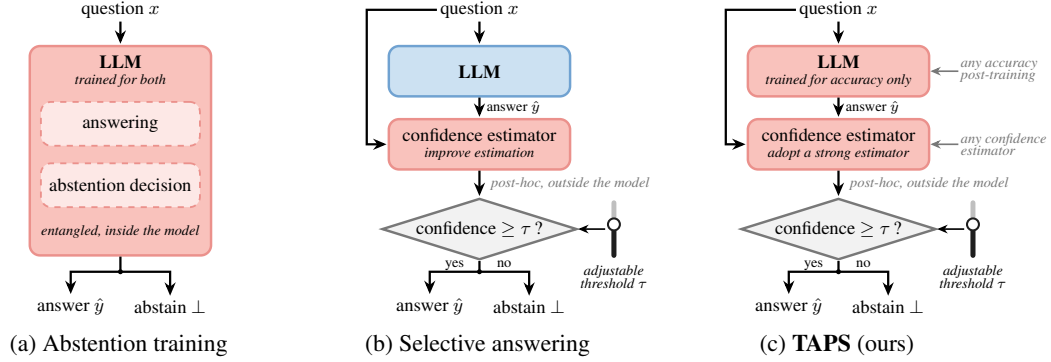
\begin{figure*}[t]
    \centering
    \input{figures/method_demo}
    \caption{
\textbf{Comparison of abstention approaches.}
\subref*{fig:method-demo-abstention-training} Abstention training jointly optimizes answering and abstention decisions inside the model.
\subref*{fig:method-demo-selective-answering} Selective answering externalizes the abstention decision after generation using a confidence estimator and an adjustable threshold $\tau$.
\subref*{fig:method-demo-taps} TAPS trains the answering model solely for accuracy while retaining the same post-hoc selection procedure.}
    \label{fig:method-demo}
\end{figure*}

\Cref{sec:motivation} shows that stronger confidence estimation largely closes the gap in distinguishing ability, while leaving the gap in answering ability unaddressed.
Since post-training for task performance is already standard practice, a natural remedy is to apply selective answering on top of a post-trained model.
We refer to this recipe as \textbf{T}rain for \textbf{A}ccuracy, then \textbf{P}ost-hoc \textbf{S}elect (\textbf{TAPS}), as illustrated in \Cref{fig:method-demo-taps}.
TAPS is an enhancement of selective answering (\Cref{fig:method-demo-selective-answering}).
Unlike abstention training (\Cref{fig:method-demo-abstention-training}), it separates answering and rejection, keeps the abstention decision outside the model, and therefore retains controllability over coverage.
Specifically, TAPS consists of two stages.

\paragraph{Stage 1: Accuracy training.}
Post-training for task accuracy has become a standard practice, with established methods directly applicable to this stage \citep{deepseek2025r1, yu2025dapo}.
Following common practice, we instantiate TAPS with outcome-reward RL:
\begin{equation}
\label{eq:ans_reward}
r_{\mathrm{ans}}(x,\hat y)
=
\begin{cases}
+1, & \text{if } \hat y \text{ is a valid answer judged correct}, \\
-1, & \text{otherwise},
\end{cases}
\end{equation}
We optimize this reward using the widely used GRPO \citep{shao2024deepseekmath} algorithm.
Starting from \textit{PreRL}, \ie, the model before accuracy training, we obtain the trained model denoted \textit{AccRL}.

\paragraph{Stage 2: Post-hoc selection.}
At inference time, the AccRL model generates a candidate answer $\hat y$ for a question $x$, and an external confidence estimator assigns a score $s(x,\hat y)$, with higher scores indicating greater confidence.
Given an acceptance threshold $\tau$, the selective answering function is
\begin{equation}
\label{eq:taps_selection}
f_\tau(x)
=
\begin{cases}
\hat y, & \text{if } s(x,\hat y) \geq \tau, \\
\bot,   & \text{otherwise},
\end{cases}
\end{equation}
where $\bot$ denotes abstention.
This stage can directly reuse existing confidence estimation methods, including unsupervised approaches such as semantic entropy \citep{semantic_entropy}, supervised approaches such as probing \citep{orgrad-2025-llmsknow}, or direct calibration training \citep{damani2026beyond-binary-reward}.

Beyond serving as a controlled comparison, this construction has two broader implications.
\begin{enumerate*}[label=\roman*)]
    \item[\ding{182}]
    Given that TAPS, a straightforward extension of selective answering, achieves competitive performance, the additional benefits of internalizing abstention decisions need to be carefully evaluated.
    \item[\ding{183}]
    TAPS is also composed of two independently well-studied components: standard post-training and well-developed confidence estimation, so advances in either can be incorporated directly, making it an evolving baseline against which abstention training should be continually reassessed.
\end{enumerate*}

%% file: figures/method_demo.tex
\begingroup
\definecolor{frozenfill}{RGB}{204,225,242}
\definecolor{frozenline}{RGB}{112,162,209}
\definecolor{trainfill}{RGB}{249,194,188}
\definecolor{trainline}{RGB}{235,145,140}
\definecolor{posthocfill}{RGB}{242,242,242}
\definecolor{posthocline}{RGB}{120,120,120}
\definecolor{knobcol}{RGB}{30,30,30}

\newcommand{\boxtag}[1]{\\[-2pt]{\scriptsize\itshape #1}}

\tikzset{
  flow/.style={-{Stealth[length=2.3mm,width=1.9mm]}, line width=1.1pt, shorten >=0.5pt},
  thinflow/.style={-{Stealth[length=1.8mm,width=1.5mm]}, line width=0.8pt},
  blk/.style={draw, rounded corners=5pt, line width=1.2pt, align=center,
              inner sep=2pt, minimum width=3.1cm, minimum height=0.86cm, font=\small},
  frozen/.style ={blk, fill=frozenfill,  draw=frozenline},
  trained/.style={blk, fill=trainfill,   draw=trainline},
  posthoc/.style={blk, fill=posthocfill, draw=posthocline},
  estimator/.style={blk, fill=trainfill, draw=trainline},
  innerdecision/.style={blk, fill=trainfill!38,
              draw=trainline, dashed, line width=0.9pt, minimum width=2.72cm},
  decision/.style={diamond, aspect=3.3, draw=posthocline, fill=posthocfill,
              line width=1.2pt, align=center, inner sep=1pt,
              minimum width=3.1cm, minimum height=1.05cm, font=\footnotesize},
  sidenote/.style={font=\scriptsize\itshape, text=posthocline,
              align=left, inner sep=1pt},
  io/.style={font=\small, inner sep=1.5pt, align=center},
  edgelab/.style={font=\scriptsize, inner sep=1.5pt},
  knoblabel/.style={font=\scriptsize\itshape, align=center, inner sep=1pt},
}

\newcommand{\tauslider}[1]{%
  \begin{scope}[shift={(#1)}]
    \draw[line width=2.4pt, line cap=round, black!25] (0,-0.45) -- (0,0.45);
    \draw[line width=2.4pt, line cap=round, knobcol]  (0,-0.45) -- (0,0.08);
    \filldraw[fill=white, draw=knobcol, line width=1.0pt] (0,0.08) circle (2.9pt);
  \end{scope}}

\def\yQ{0}       
\def\yLLM{-1.05} \def\yCE{-2.30} \def\yTH{-3.75} 
\def\yOUT{-5.08} 
\def\dx{0.85}    
\def\xK{2.30}    

\captionsetup[subfigure]{labelformat=simple,hypcap=true,
    font=footnotesize,margin={0pt,1.5cm}}
\renewcommand{\thesubfigure}{(\alph{subfigure})}
\begin{subfigure}[t]{0.32\linewidth}
    \centering
    \resizebox{!}{4.2cm}{%
        \begin{tikzpicture}[x=1cm,y=1cm]
            \node[io] (qa) at (0,\yQ) {question $x$};
            \node[trained, minimum height=3.58cm] (boxa) at (0,-2.41) {};
            \node[align=center, font=\small] (lla) at (0,\yLLM) {\textbf{LLM}\boxtag{trained for both}};
            \node[innerdecision, minimum height=0.75cm] (ansa) at (0,-1.95) {answering};
            \node[innerdecision, minimum height=0.75cm] (dea) at (0,-3.00) {abstention decision};
            \node[font=\scriptsize\itshape, align=center] at (0,-3.80) {entangled, inside the model};
            \draw[flow] (qa) -- (boxa.north);
            \node[io] (oa1) at (-\dx,\yOUT) {answer $\hat{y}$};
            \node[io] (oa2) at ( \dx,\yOUT) {abstain $\bot$};
            \coordinate (splita) at (0,-4.48);
            \draw[line width=1.1pt] (boxa.south) -- (splita);
            \fill (splita) circle (1.2pt);
            \draw[flow] (splita) -| (oa1.north);
            \draw[flow] (splita) -| (oa2.north);
            \pgfresetboundingbox
            \path[use as bounding box] (-1.90,-5.25) rectangle (3.80,0.18);
        \end{tikzpicture}%
    }
    \caption{Abstention training}
    \label{fig:method-demo-abstention-training}
\end{subfigure}%
\hfill
\begin{subfigure}[t]{0.32\linewidth}
    \centering
    \resizebox{!}{4.2cm}{%
        \begin{tikzpicture}[x=1cm,y=1cm]
            \node[io] (qb) at (0,\yQ) {question $x$};
            \node[frozen]  (llmb) at (0,\yLLM) {\textbf{LLM}};
            \node[estimator] (ceb) at (0,\yCE) {confidence estimator\boxtag{improve estimation}};
            \node[decision] (thb) at (0,\yTH) {confidence $\geq \tau$\,?};
            \node[sidenote, anchor=west] at (0.16,-2.98) {post-hoc, outside the model};
            \draw[flow] (qb) -- (llmb);
            \draw[flow] (llmb) -- node[edgelab,right=1.5pt] {answer $\hat{y}$} (ceb);
            \draw[flow] (qb.west) -- ++(-1.2,0) |- (ceb.west);
            \draw[flow] (ceb) -- (thb);
            \node[io] (ob1) at (-\dx,\yOUT) {answer $\hat{y}$};
            \node[io] (ob2) at ( \dx,\yOUT) {abstain $\bot$};
            \coordinate (splitb) at (0,-4.48);
            \draw[line width=1.1pt] (thb.south) -- (splitb);
            \fill (splitb) circle (1.2pt);
            \draw[flow] (splitb) -- (-\dx,-4.48) -- (ob1.north);
            \draw[flow] (splitb) -- ( \dx,-4.48) -- (ob2.north);
            \node[edgelab, anchor=base] at (-0.5*\dx,-4.39) {yes};
            \node[edgelab, anchor=base] at ( 0.5*\dx,-4.39) {no};
            \tauslider{\xK,\yTH}
            \draw[thinflow] (\xK-0.15,\yTH) -- (thb.east);
            \node[knoblabel, anchor=north] at (\xK,\yTH-0.58) {adjustable\\[-2pt]threshold $\tau$};
            \pgfresetboundingbox
            \path[use as bounding box] (-1.90,-5.25) rectangle (3.80,0.18);
        \end{tikzpicture}%
    }
    \caption{Selective answering}
    \label{fig:method-demo-selective-answering}
\end{subfigure}%
\hfill
\begin{subfigure}[t]{0.32\linewidth}
    \centering
    \resizebox{!}{4.2cm}{%
        \begin{tikzpicture}[x=1cm,y=1cm]
            \node[io] (qc) at (0,\yQ) {question $x$};
            \node[trained] (llmc) at (0,\yLLM) {\textbf{LLM}\boxtag{trained for accuracy only}};
            \node[estimator] (cec) at (0,\yCE) {confidence estimator\boxtag{adopt a strong estimator}};
            \node[sidenote, anchor=west] (accuracyhint) at (2.05,\yLLM) {any accuracy\\[-2pt]post-training};
            \node[sidenote, anchor=west] (confidencehint) at (2.05,\yCE) {any confidence\\[-2pt]estimator};
            \draw[thinflow, posthocline] (accuracyhint.west) -- (llmc.east);
            \draw[thinflow, posthocline] (confidencehint.west) -- (cec.east);
            \node[decision] (thc) at (0,\yTH) {confidence $\geq \tau$\,?};
            \node[sidenote, anchor=west] at (0.16,-2.98) {post-hoc, outside the model};
            \draw[flow] (qc) -- (llmc);
            \draw[flow] (llmc) -- node[edgelab,right=1.5pt] {answer $\hat{y}$} (cec);
            \draw[flow] (qc.west) -- ++(-1.2,0) |- (cec.west);
            \draw[flow] (cec) -- (thc);
            \node[io] (oc1) at (-\dx,\yOUT) {answer $\hat{y}$};
            \node[io] (oc2) at ( \dx,\yOUT) {abstain $\bot$};
            \coordinate (splitc) at (0,-4.48);
            \draw[line width=1.1pt] (thc.south) -- (splitc);
            \fill (splitc) circle (1.2pt);
            \draw[flow] (splitc) -- (-\dx,-4.48) -- (oc1.north);
            \draw[flow] (splitc) -- ( \dx,-4.48) -- (oc2.north);
            \node[edgelab, anchor=base] at (-0.5*\dx,-4.39) {yes};
            \node[edgelab, anchor=base] at ( 0.5*\dx,-4.39) {no};
            \tauslider{\xK,\yTH}
            \draw[thinflow] (\xK-0.15,\yTH) -- (thc.east);
            \node[knoblabel, anchor=north] at (\xK,\yTH-0.58) {adjustable\\[-2pt]threshold $\tau$};
            \pgfresetboundingbox
            \path[use as bounding box] (-1.90,-5.25) rectangle (3.80,0.18);
        \end{tikzpicture}%
    }
    \caption{\textbf{TAPS} (ours)}
    \label{fig:method-demo-taps}
\end{subfigure}%
\endgroup

%% file: sections/05_experiments.tex
\clearpage

\section{Experiments}
\label{sec:experiments}

\input{tables/RQ1_main_table}

We organize the empirical study around three questions:
\begin{itemize}[leftmargin=*, topsep=1pt, itemsep=1pt]
    \item \textbf{RQ1:} Does accuracy training make TAPS competitive with abstention training?
    \item \textbf{RQ2:} How does accuracy training affect the two abilities, and thus abstention performance?
    \item \textbf{RQ3:} How does post-hoc selection affect distinguishing ability, and thus abstention performance?
\end{itemize}

\subsection{Experimental Setup}
\label{sec:experimental-setup}

\paragraph{Models and datasets.}
Following recent abstention studies \citep{wei2025truthrl,zeng2026ucpo}, we evaluate two instruction-tuned models, Qwen2.5-7B-Instruct \citep{qwen25technicalreport} and Llama-3.1-8B-Instruct \citep{grattafiori2024llama3}.
Given the growing adoption of chain-of-thought (CoT) reasoning, we also include a reasoning model, Qwen3-8B \citep{yang2025qwen3}, and elicit CoT from the instruction-tuned models via prompts (provided in Appendix~\ref{app:prompts}).
To evaluate factual hallucination, we use three factual QA benchmarks that span a range of knowledge types and difficulty levels: Natural Questions (NQ) \citep{kwiatkowski2019natural}, PopQA \citep{mallen-etal-2023-trust}, and TriviaQA \citep{joshi-etal-2017-triviaqa}.
All experiments use a closed-book setting.
Dataset splits are provided in Appendix~\ref{app:data-splits}.

\paragraph{Compared methods.}
We compare TAPS with the representative SFT-based R-Tuning \citep{zhang-etal-2024-r} and the state-of-the-art RL method TruthRL \citep{wei2025truthrl}.
We also include UCPO \citep{zeng2026ucpo}, a recent RL method that addresses the behavior collapse problem through dynamic rewards.
For TruthRL, we test three abstention rewards: $r_{\mathrm{abs}}\in\{-0.5,0,0.5\}$.
More details of baselines are given in Appendix~\ref{app:abstention-methods}.
For TAPS, we instantiate Stage~2 with four representative confidence estimators: \textit{Vanilla} mean log-probability, self-evaluation-based \textit{P(True)} \citep{kadavath2022language}, sampling-based \textit{semantic entropy (S.E.)} \citep{semantic_entropy,farquhar2024detecting}, and a supervised correctness \textit{probe} \citep{orgrad-2025-llmsknow}.
Implementation details are provided in Appendix~\ref{app:confidence-estimation}.

\paragraph{Training and evaluation.}
We implement accuracy training with the \texttt{verl} framework \citep{sheng-2025-hybridflow}, using the widely adopted GRPO algorithm \citep{shao2024deepseekmath}.
Hyperparameters and prompts are provided in Appendices~\ref{app:training-configuration} and~\ref{app:prompts}, respectively.
For evaluation, we greedily decode one candidate answer per question.
Qwen2.5-72B-Instruct \citep{qwen25technicalreport} assesses correctness during training, while gpt-oss-20b \citep{agarwal2025gpt-oss} independently evaluates test outputs, which avoids exploitation of the training judge.
We further analyze judge reliability in Appendix~\ref{app:llm-judge-reliability}.

\paragraph{Metrics.}
Following \Cref{sec:preliminaries}, for each abstention-trained method $m$ with native coverage $C_m$, we threshold TAPS to match $C_m$ and report Risk@$C_m$.
For curve-level analyses, we report AURC.
We additionally use answer accuracy and AUROC as diagnostic proxies for answering and distinguishing abilities, respectively.
Full metric definitions and computation details are provided in Appendix~\ref{app:evaluation}.

\subsection{Comparison with Abstention Training (RQ1)}
\label{sec:rq1}

\begin{figure*}[t]
    \centering
    \includegraphics[width=\linewidth]{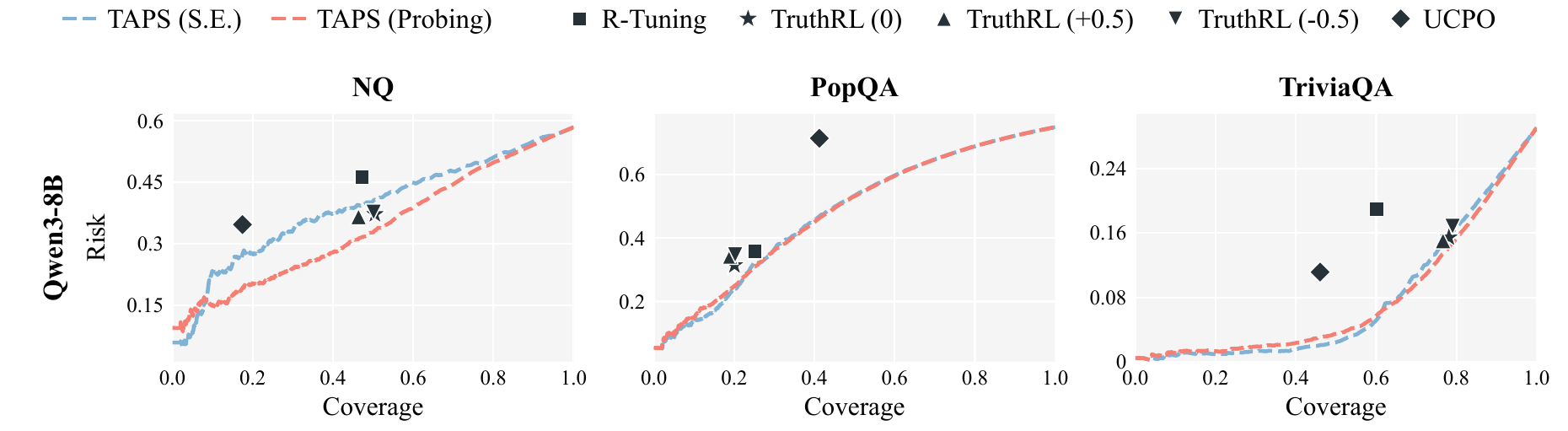}
    \caption{
    Risk--coverage comparison with abstention training.
    Each curve is obtained by varying the acceptance threshold $\tau$ for a fixed answering checkpoint and confidence estimator.
    Results for the other two models are reported in Appendix~\ref{app:extended-results-for-risk-coverage-curve}.
    }
    \label{fig:rq1-risk-coverage}
\end{figure*}

We first assess whether accuracy training improves abstention performance, and whether this improvement makes TAPS competitive with strong abstention training baselines while retaining controllability.
\Cref{tab:rq1-main-table} shows the main results, and \Cref{fig:rq1-risk-coverage} visualizes controllability through risk--coverage curves.

\paragraph{Accuracy training strengthens selective answering.}
\Cref{tab:rq1-main-table} isolates the contribution of accuracy training by comparing TAPS with its selective answering counterpart at matched coverage.
Across 90 matched-coverage comparisons, accuracy training lowers risk in 81 cases, by 3.55 pp on average, with TAPS (Probing) consistently achieving lower risk.
These results show that improving the underlying answerer directly strengthens abstention performance beyond post-hoc selection alone.

\paragraph{TAPS is competitive with abstention training.}
Against R-Tuning and UCPO, selective answering with S.E. or probing already attains lower mean risk.
Adding accuracy training generally lowers this risk further.
The one exception is S.E. at UCPO's coverage, where the mean risk increases slightly (24.63\% to 25.53\%).
Even so, it remains well below that of UCPO (31.96\%).
Against the stronger TruthRL baseline, TAPS with probing achieves lower risk in 17 of 27 comparisons (mean advantage 1.49 pp) across its three abstention rewards, while TAPS with S.E. also remains comparable on average (mean advantage 0.17 pp).
Accuracy training thus makes TAPS a strong competitor to abstention training.
Notably, TAPS shows a stronger advantage on the reasoning model: On Qwen3-8B, TAPS with probing and S.E. reduce mean risk relative to TruthRL by 5.05 and 2.40 pp, respectively, whereas their mean risks are higher on the two instruction-tuned models (0.30 and 0.94 pp).
This advantage makes TAPS particularly promising as reasoning models see wider deployment.

\paragraph{TAPS retains direct control over coverage.}
\Cref{fig:rq1-risk-coverage} shows the controllability advantage of TAPS on Qwen3-8B.
Results for the other two models appear in Appendix~\ref{app:extended-results-for-risk-coverage-curve}.
For TAPS, varying the acceptance threshold traces a full risk--coverage curve, letting one trained model serve different coverage requirements.
Abstention-trained policies, in contrast, offer far weaker control: each checkpoint yields a single operating point, and even across TruthRL's three separately trained reward variants, coverage spans less than 5 percentage points in 8/9 settings.
Moreover, a lower abstention reward does not necessarily increase coverage.
One might instead ask whether a post-hoc selector could be applied to abstention-trained models to also enable coverage control.
However, this conflicts with the trained model's internalized abstention behavior, so the selector can only trace a partial risk--coverage curve below its native coverage.
We discuss this problem in detail in Appendix~\ref{app:truthrl-posthoc-selection}.

\subsection{Effects of Accuracy Training (RQ2)}
\label{sec:rq2}

\begin{figure*}[t]
    \centering
    \includegraphics[width=0.95\linewidth]{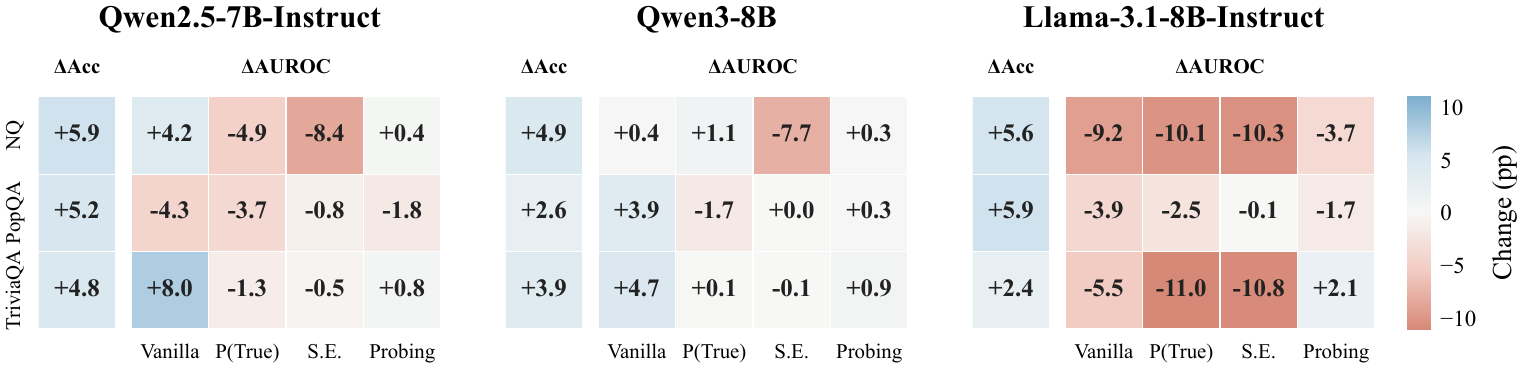}
    \caption{
    Effects of accuracy training on answering ability (Acc) and distinguishing ability (AUROC).
    Values show AccRL minus PreRL in percentage points (pp).
    }
    \label{fig:rq2-heatmap}
\end{figure*}

Accuracy training strengthens selective answering, but does it do so by improving answering ability, distinguishing ability, or both?
We next study how these two abilities change during accuracy training and how their combined effects translate into abstention performance, using Acc and AUROC as proxy metrics for these abilities and AURC for overall performance as specified in \Cref{sec:experimental-setup}.

\paragraph{Accuracy training consistently improves answering ability, but its effect on distinguishing ability varies.}
\Cref{fig:rq2-heatmap} shows that accuracy RL improves Acc across all settings, by 2.4--5.9 pp, while its effect on AUROC is far less uniform: most changes are modest for the two Qwen models, whereas unsupervised estimators (Vanilla, P(True), and S.E.) often lose substantial AUROC for Llama-3.1-8B-Instruct, especially on NQ and TriviaQA.
Probing, in contrast, remains comparatively stable, consistent with its being refit on each checkpoint's representations, which lets the selector adapt after accuracy training.
This stability aligns with the strong performance of probing in \Cref{sec:rq1}.

\begin{wrapfigure}[12]{r}{0.35\columnwidth}
    \vspace{-12pt}
    \centering
    \includegraphics[width=\linewidth]{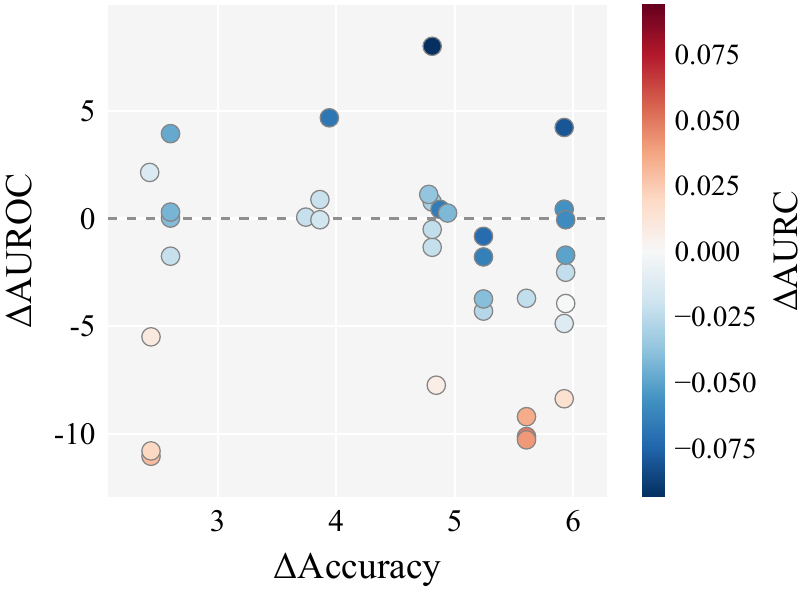}
    \captionsetup{skip=3pt}
    \caption{
    AURC changes from PreRL to AccRL.
    }
    \label{fig:rq2-aurc-comparison}
\end{wrapfigure}

\paragraph{Answering gains can outweigh degraded distinguishing ability.}
\Cref{fig:rq2-aurc-comparison} relates the changes in Acc and AUROC to the resulting change in AURC for each model--dataset--estimator configuration.
Overall, accuracy training reduces AURC in 28 of the 36 configurations, consistent with the matched-coverage risk reduction observed in \Cref{sec:rq1}.
A closer look gives an interesting result: many configurations achieve lower AURC despite degraded AUROC, showing that stronger answering ability can improve the overall trade-off even as confidence ranking deteriorates.
But this is not universal, as configurations with larger ranking degradation can still exhibit worse AURC.
Therefore, neither improved accuracy nor degraded AUROC alone determines abstention performance.
Whether answering gains translate into better performance depends on how much distinguishing ability is preserved after accuracy training.

\subsection{Effects of Confidence Estimation (RQ3)}
\label{sec:rq3}

Beyond accuracy training, TAPS also depends on the confidence estimator.
We further compare how different estimators affect distinguishing ability and the resulting abstention performance.

\begin{wrapfigure}[13]{r}{0.35\columnwidth}
    \centering
    \includegraphics[width=\linewidth]{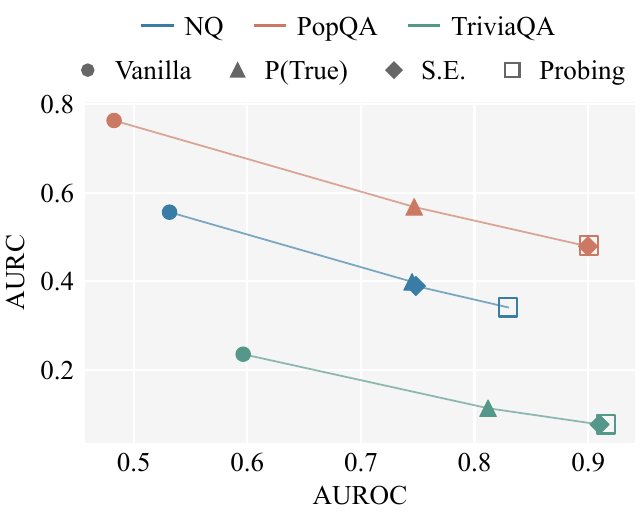}
    \captionsetup{skip=3pt}
    \caption{
    Relationship between AUROC and AURC.
    }
    \label{fig:rq3-auroc-aurc-main}
\end{wrapfigure}

\paragraph{Stronger confidence estimation generally improves abstention performance.}
Within each model--dataset setting, we fix the AccRL checkpoint and the greedy decoding configuration and vary only the estimator.
This holds the answering ability fixed, isolating the effect of confidence estimation.
\Cref{fig:rq3-auroc-aurc-main} illustrates the results on Qwen3-8B, while Appendix~\ref{app:extended-result-for-stage-2-analysis} reports results for the other models.
Across all nine model--dataset settings, higher AUROC generally corresponds to lower AURC.
P(True), S.E., and probing all achieve higher AUROC than Vanilla, and also lower AURC.
The strongest estimator remains setting-dependent: probing is more consistently strong overall, whereas S.E. performs best on PopQA and TriviaQA for Qwen2.5-7B-Instruct.
These results establish the value of strong confidence estimation and highlight the complementary roles of the two TAPS stages:
Stage~1 improves the candidate answers but can often degrade the model's own confidence signals, while Stage~2 improves upon the degraded signals to enable better abstention decisions.

\subsection{Discussion}
\label{sec:discussion}

\paragraph{Implications for abstention training.}
Our results suggest a stricter way to assess the value of learning to abstain.
Abstention training jointly changes two aspects of the model: how well it answers and how well it distinguishes wrong answers to withhold.
Consequently, improvements over the starting model conflate gains in both aspects.
Our decomposition separates these effects and shows that simply equipping selective answering with strong confidence estimators can match much of the distinguishing gain, while the remaining advantage primarily comes from improved answering.
This makes an accuracy-trained answerer paired with a strong post-hoc selector a more informative reference for identifying what abstention training contributes beyond leaving decisions post-hoc.

\paragraph{Practical considerations.}
The deployment cost of TAPS depends largely on the confidence estimator chosen for Stage~2, creating a flexible performance--cost trade-off.
Semantic entropy requires multiple sampled generations and semantic comparisons, whereas probing requires correctness-labeled responses and a one-time fitting step but adds only a lightweight classifier at inference time.
Cheaper estimators such as Vanilla confidence can instead be used when computational cost is more important.
Importantly, TAPS itself is not tied to any particular estimator: improved confidence estimators can be substituted directly without redesigning the answering model or introducing a new abstention-training objective.
This modularity allows the selection mechanism to be chosen according to the deployment setting and improved independently over time.

%% file: tables/RQ1_main_table.tex
\begin{table*}[t]
    \centering
    \caption{
    Risk (\%, $\downarrow$) at matched coverage.
    Each group compares an abstention training method with selective answering and TAPS at the same coverage (\%) shown in the shaded row.
    Avg.\ averages the nine model--dataset pairs within each group.
    For TruthRL, the value in parentheses is the abstention reward $r_{\mathrm{abs}}$.
    The lowest risk in each group is in \textbf{bold}, and the second-lowest is \underline{underlined}.
    }
    \label{tab:rq1-main-table}
    \setlength{\tabcolsep}{3pt}
    \renewcommand{\arraystretch}{1.0}

    \newcommand{\TTSBand}{\rowcolor{gray!12}}
    \newcommand{\TTSBlockSep}{\addlinespace[2pt]}
    \newcommand{\TTSCov}[1]{#1}
    \newcommand{\TTSCovOf}[1]{\rule{0pt}{2.5ex}\rule[-1.0ex]{0pt}{0pt}\textit{Coverage of} #1}
    \newcommand{\TTSz}{\phantom{0}}
    \newcommand{\TTSInd}{\hspace{0.9em}}
    \newcommand{\TTSOurs}[1]{\TTSInd TAPS (#1)}
    \newcommand{\TTSPre}[1]{\TTSInd Selective Answering (#1)}
    \newcommand{\TTSpend}{\rlap{$^{\dagger}$}}

    \begin{adjustbox}{max width=\textwidth}
    \begin{tabular}{l ccc c ccc c ccc c c}
    \toprule
    \multirow{2.5}{*}{\textbf{Risk@Cov. (\%)} $\downarrow$}
      & \multicolumn{3}{c}{\textbf{Qwen2.5-7B-Instruct}} &
      & \multicolumn{3}{c}{\textbf{Qwen3-8B}} &
      & \multicolumn{3}{c}{\textbf{Llama-3.1-8B-Instruct}} &
      & \multirow{2.5}{*}{\textbf{Avg.}} \\
    \cmidrule(lr){2-4}\cmidrule(lr){6-8}\cmidrule(lr){10-12}
      & NQ & PopQA & TriviaQA &
      & NQ & PopQA & TriviaQA &
      & NQ & PopQA & TriviaQA & & \\
    \midrule

    \TTSBand
    \TTSCovOf{R-Tuning}
      & \TTSCov{32.51} & \TTSCov{15.16} & \TTSCov{48.63} &             
      & \TTSCov{47.23} & \TTSCov{25.18} & \TTSCov{60.16} &             
      & \TTSCov{57.22} & \TTSCov{24.67} & \TTSCov{73.86} & & \\        
    \TTSInd R-Tuning
      & 41.11 & 25.60 & 12.23 &                      
      & 46.29 & 35.77 & 18.94 &                      
      & 42.85 & 25.56 & 12.46 & & 28.98 \\        
    \TTSPre{S.E.}
      & \underline{34.47} & \underline{24.52} & \TTSz 8.50 &         
      & \underline{39.14} & 35.48 & \TTSz 8.57 &         
      & 32.09 & 22.57 & \TTSz \underline{7.14} & & 23.61 \\        
    \TTSPre{Probing}
      & 38.69 & 28.31 & 10.36 &         
      & \underline{39.14} & 35.64 & \TTSz 8.90 &         
      & \underline{31.49} & 22.30 & \TTSz 8.15 & & 24.77 \\        
    \TTSOurs{S.E.}
      & 40.57 & \textbf{19.57} & \TTSz \textbf{5.31} &         
      & 39.41 & \underline{32.12} & \TTSz \textbf{5.28} &         
      & 34.80 & \textbf{18.77} & \TTSz 9.66 & & \underline{22.83} \\        
    \TTSOurs{Probing}
      & \textbf{31.06} & 25.99 & \TTSz \underline{7.60} &                 
      & \textbf{31.58} & \textbf{30.98} & \TTSz \underline{5.79} &                 
      & \textbf{28.31} & \underline{20.48} & \TTSz \textbf{5.95} & & \textbf{20.86} \\        
    \TTSBlockSep
    \cmidrule(){1-14}
    \TTSBand
    \TTSCovOf{TruthRL ($+0.5$)}
      & \TTSCov{46.95} & \TTSCov{23.24} & \TTSCov{72.66} &             
      & \TTSCov{46.37} & \TTSCov{18.99} & \TTSCov{76.66} &             
      & \TTSCov{75.48} & \TTSCov{45.51} & \TTSCov{92.36} & & \\        
    \TTSInd TruthRL ($+0.5$)
      & \underline{37.66} & \underline{29.31} & \textbf{15.21} &             
      & \underline{36.63} & 34.30 & 15.05 &             
      & \underline{34.72} & 38.78 & \textbf{11.09} & & 28.08 \\        
    \TTSPre{S.E.}
      & 43.82 & 38.75 & 22.08 &                      
      & 39.04 & 27.14 & 17.63 &                      
      & 40.14 & 39.42 & 14.56 & & 31.40 \\        
    \TTSPre{Probing}
      & 45.49 & 38.85 & 23.19 &                      
      & 38.63 & 28.83 & 17.05 &                      
      & 38.31 & 37.97 & 14.82 & & 31.46 \\        
    \TTSOurs{S.E.}
      & 42.20 & \textbf{27.90} & 17.62 &                      
      & 39.52 & \textbf{22.68} & \underline{14.74} &                      
      & 36.36 & \underline{32.72} & 13.37 & & \underline{27.46} \\        
    \TTSOurs{Probing}
      & \textbf{36.92} & 31.84 & \underline{17.61} &                      
      & \textbf{31.27} & \underline{23.99} & \textbf{13.44} &                      
      & \textbf{34.53} & \textbf{32.00} & \underline{11.96} & & \textbf{25.95} \\        
    \TTSBlockSep
    \cmidrule(){1-14}
    \TTSBand
    \TTSCovOf{TruthRL ($0$)}
      & \TTSCov{50.60} & \TTSCov{19.69} & \TTSCov{72.70} &             
      & \TTSCov{50.45} & \TTSCov{20.07} & \TTSCov{78.13} &             
      & \TTSCov{77.48} & \TTSCov{43.37} & \TTSCov{90.10} & & \\        
    \TTSInd TruthRL ($0$)
      & \underline{40.54} & 29.06 & \textbf{16.21} &                      
      & \underline{37.27} & 31.45 & \underline{15.44} &             
      & \textbf{33.60} & \textbf{27.92} & \underline{10.84} & & \underline{26.93} \\        
    \TTSPre{S.E.}
      & 46.31 & 33.23 & 22.11 &             
      & 41.06 & 28.48 & 18.75 &                      
      & 40.48 & 37.71 & 13.47 & & 31.29 \\        
    \TTSPre{Probing}
      & 47.12 & 34.31 & 23.22 &             
      & 41.31 & 29.85 & 18.08 &                      
      & 38.93 & 36.53 & 13.79 & & 31.46 \\        
    \TTSOurs{S.E.}
      & 43.31 & \textbf{23.44} & 17.66 &             
      & 40.81 & \textbf{23.84} & 15.66 &                      
      & 36.97 & 32.42 & 12.37 & & 27.39 \\        
    \TTSOurs{Probing}
      & \textbf{38.58} & \underline{28.95} & \underline{17.64} &                      
      & \textbf{33.29} & \underline{24.97} & \textbf{14.20} &                      
      & \underline{35.15} & \underline{30.81} & \textbf{10.79} & & \textbf{26.04} \\        
    \TTSBlockSep
    \cmidrule(){1-14}
    \TTSBand
    \TTSCovOf{TruthRL ($-0.5$)}
      & \TTSCov{51.65} & \TTSCov{19.48} & \TTSCov{73.93} &             
      & \TTSCov{50.13} & \TTSCov{20.25} & \TTSCov{79.07} &             
      & \TTSCov{74.21} & \TTSCov{38.23} & \TTSCov{91.66} & & \\        
    \TTSInd TruthRL ($-0.5$)
      & \underline{40.34} & \underline{27.15} & \textbf{16.38} &             
      & \underline{37.76} & 34.77 & 16.83 &             
      & \underline{34.38} & \underline{27.81} & \textbf{10.50} & & 27.33 \\        
    \TTSPre{S.E.}
      & 46.78 & 32.57 & 22.90 &                      
      & 40.88 & 28.64 & 19.35 &                      
      & 39.53 & 33.75 & 14.30 & & 30.97 \\        
    \TTSPre{Probing}
      & 47.97 & 34.08 & 24.00 &                      
      & 41.07 & 30.11 & 18.71 &                      
      & 37.81 & 32.44 & 14.47 & & 31.18 \\        
    \TTSOurs{S.E.}
      & 43.95 & \textbf{23.30} & \underline{18.28} &                      
      & 40.56 & \textbf{23.92} & \underline{16.14} &                      
      & 35.82 & \textbf{27.47} & 13.25 & & \underline{26.97} \\        
    \TTSOurs{Probing}
      & \textbf{39.32} & 28.90 & 18.38 &                      
      & \textbf{32.93} & \underline{25.10} & \textbf{14.82} &                      
      & \textbf{34.01} & 27.85 & \underline{11.57} & & \textbf{25.88} \\        
    \TTSBlockSep
    \cmidrule(){1-14}
    \TTSBand
    \TTSCovOf{UCPO}
      & \TTSCov{14.71} & \TTSCov{36.72} & \TTSCov{\TTSz 55.98} &        
      & \TTSCov{17.43} & \TTSCov{41.22} & \TTSCov{46.03} &             
      & \TTSCov{60.73} & \TTSCov{25.87} & \TTSCov{74.21} & & \\        
    \TTSInd UCPO
      & 27.38 & 61.42 & 13.74 &                      
      & 34.73 & 71.55 & 11.16 &                      
      & \underline{32.50} & 27.74 & \TTSz 7.38 & & 31.96 \\        
    \TTSPre{S.E.}
      & \textbf{19.84} & 53.05 & \TTSz 11.95 &         
      & \textbf{17.55} & 51.64 & \TTSz 3.24 &         
      & 33.52 & 23.61 & \TTSz \underline{7.26} & & \underline{24.63} \\        
    \TTSPre{Probing}
      & 28.17 & 52.99 & \TTSz 13.71 &         
      & 19.56 & 51.81 & \TTSz 5.10 &         
      & 32.56 & 23.15 & \TTSz 8.24 & & 26.14 \\        
    \TTSOurs{S.E.}
      & 36.90 & \underline{44.24} & \TTSz \textbf{8.74} &         
      & 27.42 & \underline{46.89} & \TTSz \textbf{2.08} &         
      & 34.48 & \textbf{19.38} & \TTSz 9.61 & & 25.53 \\        
    \TTSOurs{Probing}
      & \underline{23.81} & \textbf{44.17} & \TTSz \underline{9.77} &                 
      & \underline{19.20} & \textbf{46.31} & \TTSz \underline{3.14} &                 
      & \textbf{29.61} & \underline{20.96} & \TTSz \textbf{6.01} & & \textbf{22.55} \\        

    \bottomrule
    \end{tabular}
    \end{adjustbox}
\end{table*}

%% file: sections/07_conclusion.tex
\section{Conclusion}
\label{sec:conclusion}

In this paper, we revisited LLM abstention by decomposing matched-coverage risk into answering and distinguishing ability.
The decomposition shows that abstention training's advantage over strong selective answering comes from better answering ability, which motivates TAPS: train for accuracy, then select post hoc.
With a probing-based estimator, TAPS obtains lower mean risk than three representative abstention training methods at matched coverage while keeping coverage adjustable.
More broadly, our results call for a more careful interpretation of progress in abstention training: better overall performance does not necessarily imply better abstention, and TAPS provides a simple way to disentangle the two.

For future work, several directions remain open.
An important next step is to test whether TAPS scales to larger models and generalizes to other tasks, including retrieval-augmented QA and agentic tasks.
Because accuracy training can weaken the model's own confidence signals, and TAPS benefits from refitting a supervised confidence estimator to recover the distinguishing ability, calibration-preserving post-training could make TAPS more efficient and robust.
Finally, developing methods that enable abstention-trained models to adjust their coverage after training is another promising direction.

%% file: sections/statements.tex
\subsection*{AI Use Statement}
In this work, we used generative AI tools for implementing our methods, assisting with translation, creating or modifying scientific figures or images, summarizing or analyzing existing literature, and editing the paper to improve readability.
We have not used generative AI tools for helping develop theoretical models or conceptual frameworks, formulating mathematical claims, providing critical ingredients for proving mathematical claims, assisting in the writing of proofs, proposing or refining hypotheses, designing or providing feedback on research methodology or experiments, or interpreting results.
The following categories are not applicable to this work: generating synthetic datasets, cleaning and reformatting datasets, and supporting qualitative and thematic data analysis.
We have reviewed all AI-assisted work. We manually verified every citation, AI-generated code, and experimental detail.
All reported numbers were reproduced from the project artifacts and independently checked by the authors. We take responsibility for the final content of this work, including text, claims, and artifacts produced with the aid of generative AI.

\subsection*{Ethics statement}

This work aims to improve the reliability and controllability of LLM abstention using publicly available models and factual QA benchmarks. Abstention does not eliminate factual errors, and confidence estimates may reflect biases in the underlying models and data. Our results should therefore not be interpreted as guarantees of correctness or safety.


\subsection*{Reproducibility statement}

\Cref{sec:experimental-setup} summarizes the experimental setup.
Appendix~\ref{app:experimental-details} details dataset splits, baseline implementations, confidence estimators, training hyperparameters, hardware, and evaluation protocols.
Appendix~\ref{app:llm-judge-reliability} reports the reliability analysis of the evaluation judge, and Appendix~\ref{app:prompts} provides the generation and judging prompts.
We will publicly release our code, including training scripts, evaluation pipelines, and prompts, upon acceptance.

%% file: sections/appendix.tex
\clearpage

\section{Selective Prediction in Classification}
\label{app:selective-prediction}
Selective prediction improves reliability by allowing a predictor to withhold uncertain outputs \citep{chow1970optimum, elyaniv2010foundations, geifman2017selective}.
Prior work broadly follows two paradigms, distinguished by whether rejection is learned jointly with prediction.
\emph{Post-hoc selection} keeps the predictor fixed and accepts predictions whose confidence exceeds a threshold \citep{geifman2017selective, geifman2018biasreduced}.
\emph{Learned rejection} instead jointly optimizes prediction and rejection within the classifier \citep{bartlett2008classification, cortes2016learning,
geifman-2019-selectivenet}.
Modern LLM abstention inherits the same distinction:
selective answering externalizes the abstention decision, whereas abstention training internalizes it.
We find that the apparent advantage of abstention training largely reflects stronger answering ability rather than more accurate abstention decisions, paralleling the observation of \citet{feng2023towards} in selective prediction that learned-rejection gains in classification can arise from a stronger classifier rather than a better rejection mechanism.

\section{Limited Coverage Control through Abstention-Reward Sweeps}
\label{app:coverage-control}
\begin{figure*}[t]
\centering  
\includegraphics[width=\textwidth]{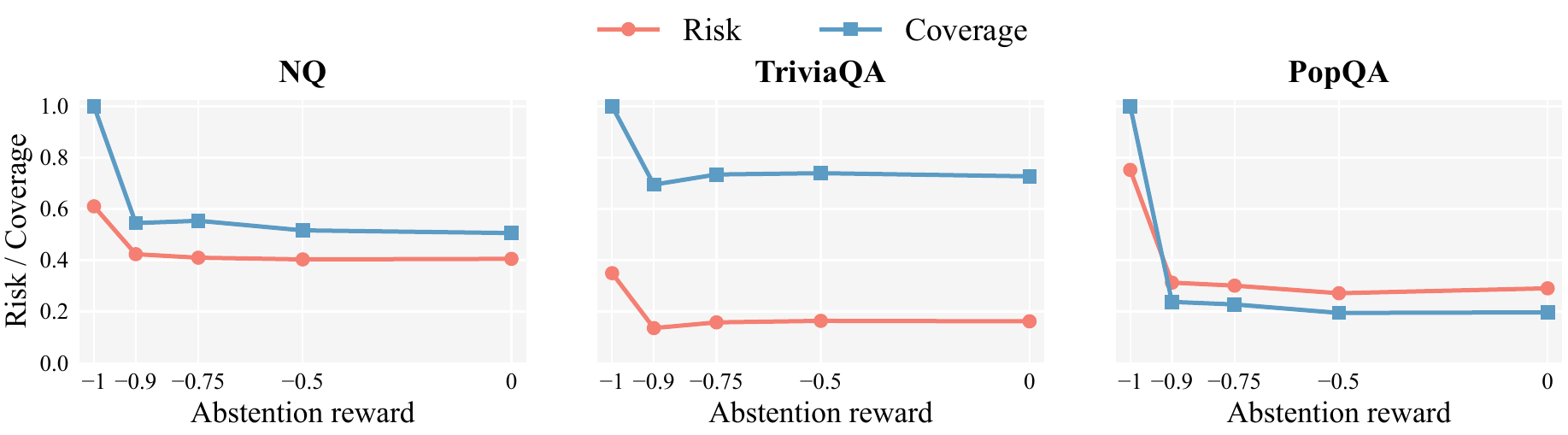}    \caption{
    \textbf{Effect of sweeping the training-time abstention reward.}
    Each point represents a separately trained TruthRL policy based on Qwen2.5-7B-Instruct across NQ, PopQA, and TriviaQA.
    A common pattern emerges: moving \(r_{\mathrm{abs}}\) from \(-1\) to \(-0.9\) causes an abrupt drop in coverage, while subsequent changes produce only limited or non-monotonic coverage variation. }
    \label{fig:abstention-reward-control}
\end{figure*}

The ternary reward in \Cref{eq:setup-truthrl-reward} makes the abstention reward \(r_{\mathrm{abs}}\) a natural training-time control signal.
Increasing \(r_{\mathrm{abs}}\) makes abstention more favorable relative to
incorrect answering and might therefore be expected to reduce coverage.
We conduct a preliminary study to examine whether varying this reward can produce policies spanning a controllable range of coverage levels.

We train separate TruthRL policies with \(
r_{\mathrm{abs}}\in\{-1,-0.9,-0.75,-0.5,0\},\) while keeping the rewards for correct and incorrect answers fixed at \(+1\) and \(-1\), respectively. Each value corresponds to a separate training run.
\Cref{fig:abstention-reward-control} shows a sharp behavioral change in coverage and risk as the abstention reward is swept. 
At \(r_{\mathrm{abs}}=-1\), abstention and incorrect answering receive the same reward, and the learned policies retain nearly full coverage. 
Increasing the reward slightly to \(r_{\mathrm{abs}}=-0.9\) triggers an
abrupt transition to substantially lower coverage.
After this transition, further increases in \(r_{\mathrm{abs}}\) do not produce a sequence of progressively lower or well-separated coverage and risk levels. Instead, the resulting policies collapse within a relatively narrow range across all datasets. 
In the risk--coverage plane, this collapse appears as a cluster of operating points within a narrow range, preventing fine-grained coverage control.

\clearpage

\section{Experimental Details}
\label{app:experimental-details}

\subsection{Data Splits}
\label{app:data-splits}

\input{tables/Appendix_data}

We evaluate different approaches across three factual QA benchmarks: Natural Questions (NQ), PopQA and TriviaQA.
For NQ and TriviaQA, whose original training splits each exceed 80,000 examples, we randomly sample 10,000 examples for training and reserve a small held-out portion for validation.
Following common practice, we adopt their validation splits as test sets, since NQ lacks an official test set and the test annotations of TriviaQA are not publicly available.
For PopQA, which provides only a single evaluation set, we randomly partition it into training, validation, and test subsets.
The final statistics for all dataset splits are summarized in \Cref{tab:setup-data-splits}.
Training is performed separately for every model--dataset pair, using only that dataset's training partition.

\subsection{Abstention Training Methods}
\label{app:abstention-methods}

\paragraph{R-Tuning.}
R-Tuning \citep{zhang-etal-2024-r} learns answer--certainty responses through SFT.
We first generate answers from the starting model on the training questions and label their correctness using the judge in Appendix~\ref{app:evaluation}.
Each supervised target contains the reference answer, followed by a certainty question and either \texttt{I am sure.} or \texttt{I am unsure.}, depending on whether the starting model answered correctly.
Questions answered incorrectly remain in the training set, but their incorrect generated answers are replaced with the reference answers in the supervised targets.
At evaluation, following their standard interpretation, we treat unsure declarations as abstentions.

\paragraph{TruthRL.}
TruthRL \citep{wei2025truthrl} uses RL to distinguish correct answering, incorrect answering, and abstention through a ternary reward.
The GRPO reward implementation uses
\begin{equation}
    r_{\mathrm{TruthRL}}(x,\hat y)=
    \begin{cases}
        +1, & \text{if $\hat y$ is a valid correct answer},\\
        r_{\mathrm{abs}}, & \text{if $\hat y$ is a valid abstention},\\
        -1, & \text{otherwise}.
    \end{cases}
    \label{eq:setup-truthrl-reward}
\end{equation}
We evaluate $r_{\mathrm{abs}}\in\{-0.5,0,0.5\}$ in the main experiments and use $r_{\mathrm{abs}}=0$ in the preliminary analysis.
Prompts are provided in Appendix~\ref{app:prompts}.

\paragraph{UCPO.}
UCPO \citep{zeng2026ucpo} uses the same three outcome categories but replaces standard GRPO advantages with uncertainty-aware estimates.
We retain the original TAD/DURA rules with $\alpha=2$, $\lambda=0.5$, $w=1$, and $\epsilon=10^{-6}$.
The implementation uses the sample standard deviation and computes its batch-level statistic over the full rollout batch.
Groups without both correct and incorrect responses receive zero advantages.
Abstention advantages are computed dynamically rather than by normalizing a fixed zero reward.
UCPO uses the same rollout budget, abstention-enabled prompts, correctness judge, and response-length limit as the other RL conditions.

\subsection{Confidence Estimation}
\label{app:confidence-estimation}

For each answering checkpoint, we adopt the same zero-temperature greedy decoding setup for answer generation, and apply different confidence-estimation methods to its generated answers.
Self-evaluation labels and additional sampled answers are used only to construct confidence scores.
They do not replace the candidate answer or introduce majority-vote answering.

\paragraph{Vanilla mean log-probability.}
Vanilla uses the average conditional log-probability of the final-answer tokens.
We extract the answer inside the final complete \texttt{\textbackslash boxed\{\}} expression and identify the tokens whose decoded spans overlap this text, denoted by $\mathcal I_{\mathrm{ans}}$.
The confidence score is
\begin{equation}
    s_{\mathrm{Vanilla}}(x,\hat y)
    =\frac{1}{|\mathcal I_{\mathrm{ans}}|}
    \sum_{t\in\mathcal I_{\mathrm{ans}}}
    \log\pi_\theta(y_t\mid x,y_{<t}).
    \label{eq:setup-vanilla}
\end{equation}
Reasoning tokens are excluded from the average but remain part of the conditioning context.
Higher scores indicate greater confidence.

\paragraph{P(True).}
Prompting-based confidence estimation asks the model to assess the reliability of its own answers \citep{kadavath2022language,tian2023calibration,xiong-2024-can}.
We use the P(True) formulation of \citet{kadavath2022language}.
The answering checkpoint receives the question and its proposed answer in a separate prompt that asks whether the answer is true.
No reference answer is supplied.
We score the label sequences \texttt{True} and \texttt{False} and normalize their probabilities:
\begin{equation}
    s_{\mathrm{P(True)}}(x,\hat y)
    =\frac{\exp\ell_{\mathrm{True}}}
    {\exp\ell_{\mathrm{True}}+\exp\ell_{\mathrm{False}}},
    \label{eq:setup-ptrue}
\end{equation}
where $\ell_v$ is the sum of conditional token log-probabilities for label sequence $v$.

\paragraph{Semantic entropy.}
Semantic entropy measures uncertainty over answer meanings by grouping semantically equivalent responses \citep{semantic_entropy,farquhar2024detecting}.
In our implementation, we draw 10 additional responses per question from the answering checkpoint with temperature $1.0$, top-$p=0.9$, and top-$k=50$.
After extracting the final answers and combining normalized duplicates, we group answers using bidirectional semantic entailment.
Two answers are grouped together only when each entails the other as an answer to the question.
With $M$ valid samples and $n_c$ samples in cluster $c$, we compute
\begin{equation}
    p_c=\frac{n_c}{M},\qquad
    H_{\mathrm{sem}}(x)=-\sum_c p_c\log p_c,\qquad
    s_{\mathrm{SE}}(x)=-H_{\mathrm{sem}}(x).
    \label{eq:setup-semantic-entropy}
\end{equation}
We use natural logarithms and exclude malformed samples from the cluster-frequency calculation.

\paragraph{Probing.}
Probing predicts answer correctness from the answering model's hidden representations \citep{orgrad-2025-llmsknow}.
We fit an $\ell_2$-regularized logistic regression on the final normalized hidden state at the last response-content token, excluding trailing special tokens and padding.
The answering model remains frozen.
Correct answers receive positive labels, while incorrect answers receive negative labels.
We remove training questions whose normalized text appears in validation or test.
The remaining questions are partitioned into 80\% fitting and 20\% development subsets using a class-stratified split.
Duplicate questions are kept within the same subset, and feature standardization is fitted only on the fitting subset.
We search inverse regularization strengths $C\in\{0.001,0.01,0.1,1,10\}$ and use L-BFGS with at most 1,000 iterations.
The classifier with the lowest development log loss is applied to validation and test without refitting.
Its predicted probability of correctness serves as the confidence score.

\subsection{Training Configuration}
\label{app:training-configuration}

\input{tables/Appendix_training}

\paragraph{RL optimization.}
We use the verl framework \citep{sheng-2025-hybridflow} for RL training.
TAPS accuracy training and TruthRL use GRPO \citep{shao2024deepseekmath}, while UCPO uses the advantage estimator described in Appendix~\ref{app:abstention-methods}.
For Qwen2.5-7B-Instruct and Llama-3.1-8B-Instruct, we use a total response budget of 256 tokens, which is sufficient for the short reasoning traces in single-hop factual QA with negligible truncation.
For Qwen3-8B, we enable native thinking and run with a larger response budget of 2048 tokens.
All RL training experiments were conducted on a single compute node
equipped with 8 NVIDIA A100 (80 GB) GPUs. 
Four GPUs are allocated to model training, and the other four serve the judge model.
\Cref{tab:setup-rl-hyperparameters} lists the shared hyperparameters.

\paragraph{SFT optimization.}
R-Tuning updates all model parameters for four epochs with a learning rate of $10^{-5}$, a global batch size of 64, and a maximum sequence length of 1,024 tokens.
We use a micro-batch of one example per training GPU and 16 gradient-accumulation steps.
The loss covers response tokens, including EOS, and excludes prompt tokens.

\subsection{Judging and Evaluation Protocol}
\label{app:evaluation}


For fair comparison, we compute all metrics on the common subset of test questions for which every compared method produces a valid, evaluable output.
This excludes output failures such as format violations or repetitive generation.
An alternative would be to count these samples as incorrect.
However, the confidence score of an invalid answer is undefined.
Assigning them artificial confidence scores could potentially distort confidence rankings and the resulting risk--coverage curves.
Moreover, we observe that these failures occur predominantly in PreRL models, so retaining them as errors could disproportionately penalize the selective answering baselines, making our approach appear better than it is.
We therefore use the common valid subset to isolate abstention performance from output failures and ensure a consistent evaluation population across methods.

\paragraph{Correctness judging.}
To support large-scale judging and facilitate experimental reproducibility, we use an LLM-as-a-judge protocol for both training rewards and evaluation.
We deploy different judge models in these two stages to avoid hacking, with Qwen2.5-72B-Instruct for training and gpt-oss-20b for evaluation.
Both judges receive the question, reference answer, and extracted final answer, without the reasoning trace.
Each judge returns \texttt{A}, \texttt{B}, or \texttt{C} for correct, incorrect, or abstention, respectively.
The full judging prompt, including all examples, is provided in Appendix~\ref{app:prompt-judge}.
We also assess the judge's reliability, with results reported in Appendix~\ref{app:llm-judge-reliability}.

\paragraph{Matched-coverage comparison.}

Let $N$ denote the number of test questions in the common evaluation subset.
For an abstention-trained policy $m$, let $c_m$, $w_m$, and $a_m$ denote its correct-answer, incorrect-attempt, and abstention counts on these questions.
Its accepted count, coverage, and risk are
\begin{equation}
    k_m = c_m + w_m,
    \qquad
    C_m = \frac{k_m}{N},
    \qquad
    R_m = \frac{w_m}{k_m}.
\end{equation}
We select the $k_m$ highest-confidence TAPS answers, randomly sampling from answers tied at the cutoff to fill the remaining slots.


\paragraph{Curve-level comparison.}

Since selective answering methods can elicit an entire risk--coverage curve, we use AURC to compare their overall performance.
We rank the $N$ answers by decreasing confidence, using the same random tie-breaking rule, and compute
\begin{equation}
    R_k = \frac{1}{k}\sum_{i=1}^{k} e_{(i)},
    \qquad
    \mathrm{AURC} = \frac{1}{N}\sum_{k=1}^{N} R_k,
\end{equation}
where $e_{(i)}$ is the error indicator of the $i$-th ranked answer and $R_k$ is the risk at coverage $k/N$.


\paragraph{Measuring answering ability.}
We use overall answer accuracy as a complementary diagnostic of the answering ability.
For a dataset of $N$ test questions, let $e_i$ denote the error indicator of the candidate answer generated for question $i$.
Answer accuracy is
\begin{equation}
    \mathrm{Acc}
    =
    1 - \frac{1}{N}\sum_{i=1}^{N} e_i.
    \label{eq:answer-accuracy}
\end{equation}
This measures the fraction of questions answered correctly under the specified generation protocol.
Because it is computed before any answers are withheld, it characterizes the quality of the generated answers without being affected by the confidence estimator or acceptance threshold.

\paragraph{Measuring distinguishing ability.}
To characterize distinguishing ability, we evaluate whether confidence scores prioritize correct answers over incorrect ones.
Let $s_i$ denote the confidence score of the $i$-th candidate answer, with higher scores indicating greater confidence, and define
$\mathcal{I}_{+}=\{i:e_i=0\}$ and
$\mathcal{I}_{-}=\{i:e_i=1\}$.
The area under the receiver operating characteristic curve (AUROC) can be computed as
\begin{equation}
    \mathrm{AUROC}
    =
    \frac{1}{|\mathcal{I}_{+}|\,|\mathcal{I}_{-}|}
    \sum_{i\in\mathcal{I}_{+}}
    \sum_{j\in\mathcal{I}_{-}}
    \left[
        \mathbf{1}\{s_i>s_j\}
        +
        \frac{1}{2}\mathbf{1}\{s_i=s_j\}
    \right],
    \label{eq:confidence-auroc}
\end{equation}
Equivalently, AUROC is the probability that a randomly chosen correct answer receives a higher confidence score than a randomly chosen incorrect answer, with half credit for ties.
An AUROC of $1$ indicates perfect separation, while a random ranking has expected AUROC $0.5$.
Since selective answering retains higher-scoring answers, this ranking measure provides a threshold-independent proxy for distinguishing ability without requiring confidence scores to be calibrated probabilities.


\section{Reliability Analysis of LLM-as-a-Judge}
\label{app:llm-judge-reliability}

\input{tables/Appendix_judge}

To validate the reliability of gpt-oss-20b as our unified evaluation judge and mitigate potential reward-hacking concerns, we conduct a human agreement analysis alongside comparisons with frontier models, including DeepSeek-V4.1-Flash, GPT 5.5, Gemini 3.6 Flash, and Claude Sonnet 5. 
We randomly sample 300 model outputs covering PreRL, AccRL, and TruthRL(0) stages, two datasets (NQ and TriviaQA), and two model families (Qwen3 and Llama), with 25 samples per cell.

\Cref{tab:judge-reliability} shows that gpt-oss-20b achieves 91.0\% overall agreement with human annotations, competitive with frontier judges such as DeepSeek-V4.1-Flash (94.3\%) and Gemini 3.6 Flash (92.3\%).
Agreement remains consistent across PreRL, AccRL, and TruthRL, at 91.0\%, 90.0\%, and 92.0\%, respectively.
More importantly, gpt-oss-20b makes the fewest false acceptances at every stage, labeling an incorrect answer or abstention as correct only once across all 300 examples, compared with 8--21 cases for the other judges.
This single case occurs on AccRL, with none on PreRL or TruthRL.
Its remaining disagreements consist of 25 false rejections, where correct answers are labeled incorrect or as abstentions, and one incorrect--abstention confusion.
Thus, its errors overwhelmingly withhold credit from correct answers rather than grant credit to non-correct outputs.
This conservative error profile provides evidence against judge leniency as an explanation for the reported low risk and supports the reliability of our evaluation.

\section{Full Behavioral Decomposition Results}
\label{app:extended-behavior-decomposition-results}
\label{app:full-decomposition}

\input{tables/Appendix_decomposition}

\Cref{tab:app-decomposition} reports the complete decomposition across all nine model--dataset settings.
In theory, selective answering leaves the generated answers unchanged and thus has $G_{\mathrm{ans}}=0$ (small nonzero values in the table arise from run-to-run generation variation across estimators despite the same zero-temperature, greedy decoding generation setup, and are negligible), so different estimators primarily affect $G_{\mathrm{dis}}$.
On average, $G_{\mathrm{dis}}$ increases from 59.47\% with Vanilla to 74.84\% with P(True) and 84.64\% with S.E., essentially matching TruthRL's 84.29\%.
TruthRL nevertheless achieves positive $G_{\mathrm{ans}}$ in every setting, averaging 5.07\%.
These results confirm that strong post-hoc selection closes the distinguishing gap, while abstention training's remaining advantage primarily comes from improved answering ability.

\section{Extended Results for Risk-Coverage Curves (RQ1)}
\label{app:extended-results-for-risk-coverage-curve}

\begin{figure*}[t]
    \centering
    \includegraphics[width=\linewidth]{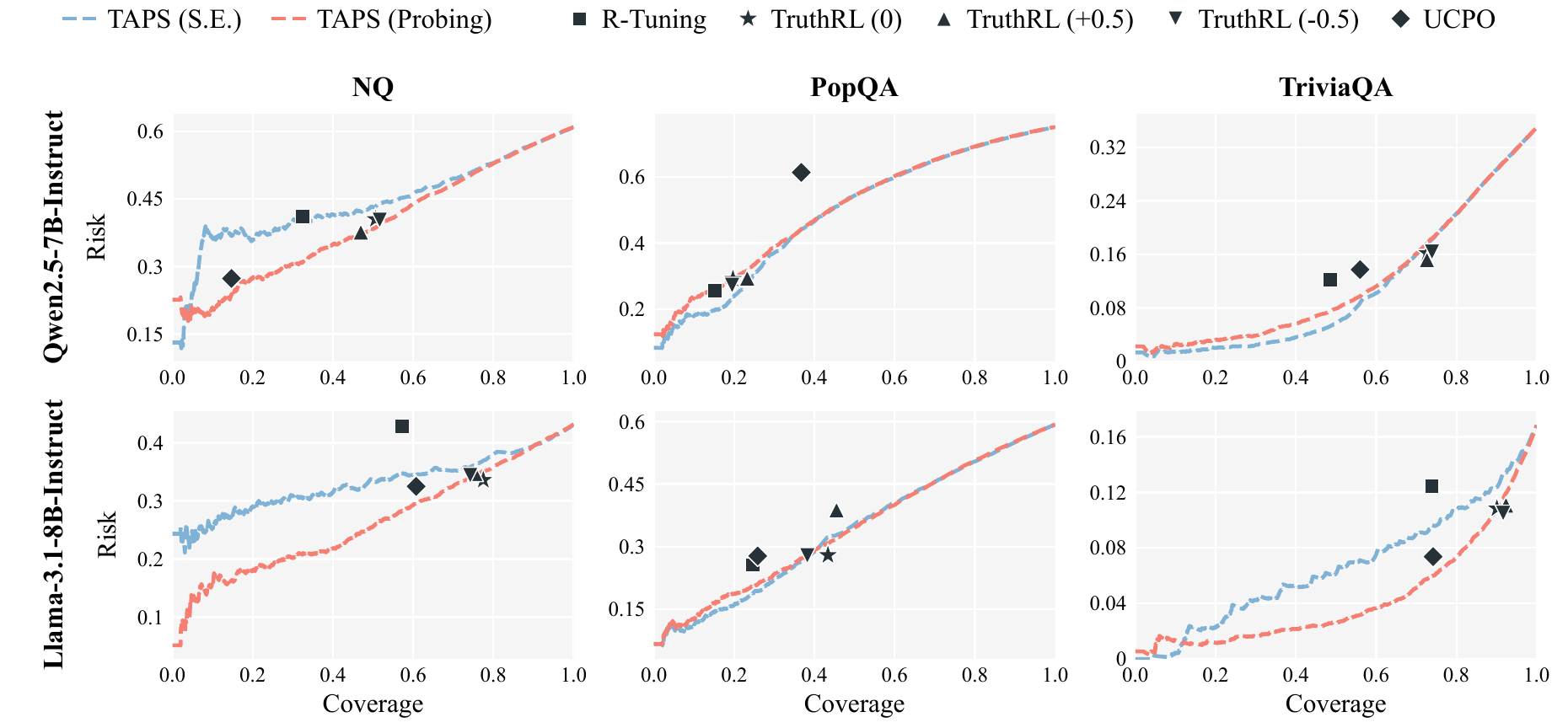}
    \caption{
    \textbf{Extended risk--coverage comparisons.}
    Results on Qwen2.5-7B-Instruct and Llama-3.1-8B-Instruct.
    TAPS curves vary the acceptance threshold.
    Markers denote the native operating points of abstention-trained checkpoints.
    }
    \label{fig:rq1-risk-coverage-appendix}
\end{figure*}

\Cref{fig:rq1-risk-coverage-appendix} extends the risk--coverage comparison to Qwen2.5-7B-Instruct and Llama-3.1-8B-Instruct.
The same pattern holds across models: varying the TAPS threshold traces a full risk--coverage curve, whereas each abstention-trained checkpoint provides only one native operating point.
Across all nine model--dataset settings, TruthRL's three reward variants span less than five percentage points of coverage in eight settings, confirming that training-time reward sweeps do not provide fine-grained coverage control.

\section{Extended Results for Stage~2 Analysis (RQ3)}
\label{app:extended-result-for-stage-2-analysis}

\begin{figure*}[t]
    \centering
    \includegraphics[width=0.7\linewidth]{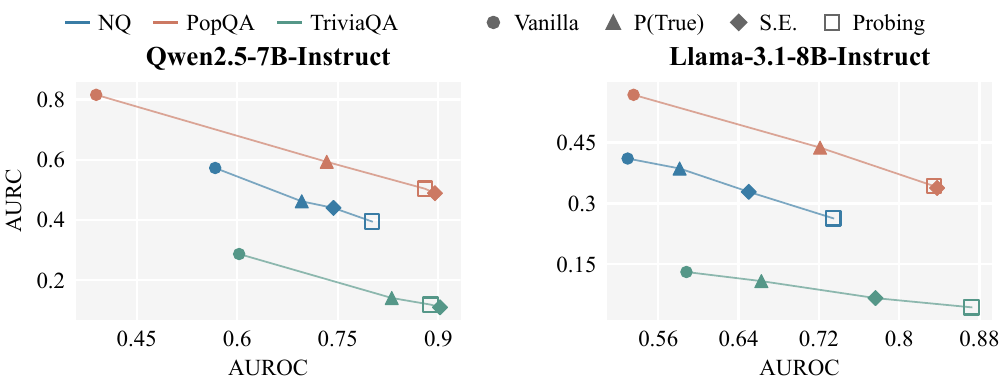}
    \caption{
    \textbf{Extended confidence-estimation results.}
    AUROC--AURC relationships on Qwen2.5-7B-Instruct and Llama-3.1-8B-Instruct.
    Each line connects confidence estimators for one dataset in increasing AUROC order.
    }
    \label{fig:rq3-auroc-aurc-appendix}
\end{figure*}

\Cref{fig:rq3-auroc-aurc-appendix} extends the Stage~2 analysis to Qwen2.5-7B-Instruct and Llama-3.1-8B-Instruct.
Within each model--dataset setting, all estimators use the same AccRL checkpoint and greedy decoding configuration, with AURC differences primarily reflecting confidence ranking.
P(True), S.E., and probing improve AUROC over Vanilla in all six settings shown, and higher AUROC generally corresponds to lower AURC.
The relationship observed on Qwen3-8B therefore generalizes across models.

\section{Diagnostic Post-hoc Selection on TruthRL}
\label{app:truthrl-posthoc-selection}

\begin{figure*}[t]
\centering  
\includegraphics[width=\textwidth]{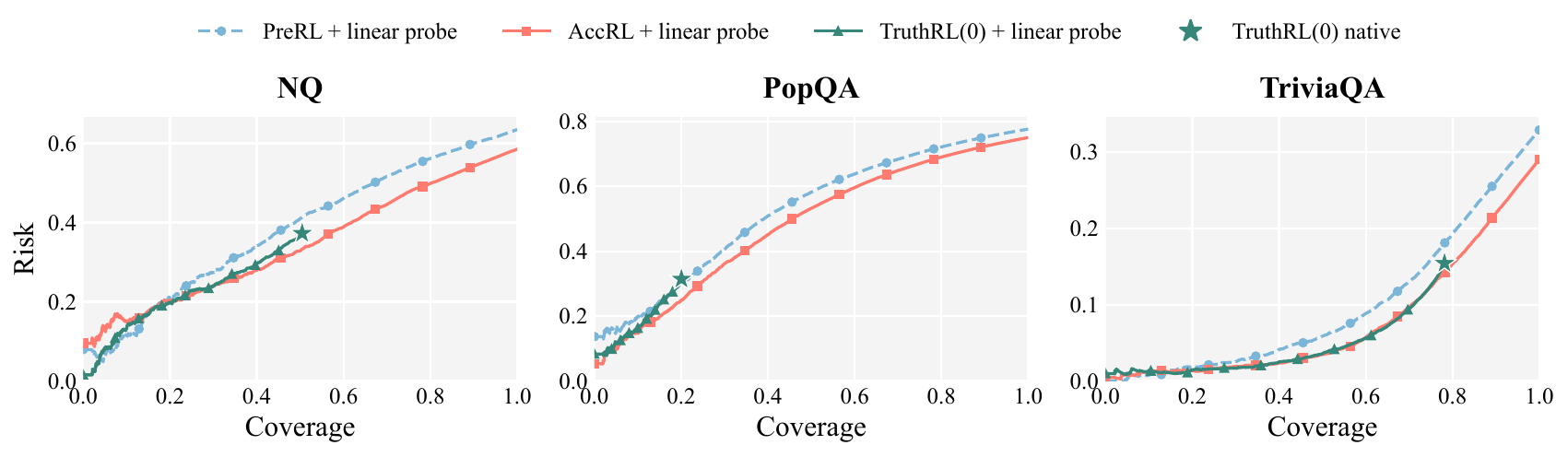}   
    \caption{
    \textbf{Post-hoc selection on TruthRL.}
    The results show Qwen3-8B across three datasets.
    Because the probe can only filter TruthRL's native answers, its curve ends at
    the native coverage (star).
    Its relative advantage over TAPS varies across datasets and coverage levels.
    }
    \label{fig:truthrl-posthoc-selection}
\end{figure*}

Abstention training and post-hoc selection embody opposing design choices:
the former internalizes the answer-or-abstain decision within the model policy, whereas the latter externalizes this decision to a confidence estimator.
Pairing TruthRL with an external selector therefore departs from the intended form of abstention training and is not a natural primary comparison.
Nevertheless, for completeness and to control for the effect of the external selector, we conduct a diagnostic analysis on Qwen3-8B across all three datasets.

A TruthRL policy produces no candidate answer for questions on which it natively abstains.
We therefore leave these abstentions unchanged and fit a separate linear probe only to the subset of questions that TruthRL(0) chooses to answer.
Sweeping the probe threshold further removes low-confidence native answers, yielding a partial risk--coverage curve over $c \in [0, c_{\mathrm{TruthRL}}]$.
In contrast, PreRL and AccRL generate a candidate answer for every question, so post-hoc selection can trace their full risk--coverage curves.

\Cref{fig:truthrl-posthoc-selection} shows no uniform ordering between TruthRL with post-hoc selection and TAPS.
On TriviaQA, their curves largely overlap throughout the shared coverage range.
On NQ and PopQA, TruthRL with probing is competitive with, and sometimes outperforms, TAPS at lower coverage, whereas TAPS performs better toward the upper end of their shared coverage range.
Thus, supplementing TruthRL with an external selector can expose additional benefits in some low-coverage regimes, but these benefits are neither uniform across datasets nor consistent across coverage levels.

Crucially, this construction can only reduce coverage from TruthRL's native operating point.
It cannot recover candidate answers for questions already rejected by the internal policy and therefore cannot access the higher-coverage portion of the curve.
This one-sided control is particularly limiting because abstention-trained policies can collapse toward narrow and conservative native coverage ranges \citep{che2026abstention,zeng2026ucpo}.
Consequently, the hybrid cannot serve as a general coverage-control mechanism comparable to post-hoc selection over an answer-generating model.
We therefore use it only as a diagnostic control for isolating the effect of
the selector.
In the main experiments, we compare the two paradigms in their intended forms: native answer-or-abstain decisions for abstention training, and adjustable post-hoc selection for TAPS.

\section{Prompts}
\label{app:prompts}

This section collects the important prompts used in our experiments.
Braced fields such as \texttt{\{question\}} are replaced with the corresponding example text.
Line wrapping and indentation are adjusted for presentation; the wording, examples, and output labels are preserved.

\paragraph{Prompts for Qwen2.5-7B-Instruct and Llama-3.1-8B-Instruct.}
\label{app:prompt-instruct}
PreRL and AccRL use the following answer-only prompt.

\begin{tcolorbox}[
  colback=gray!5,
  colframe=gray!75!black,
  title=Answer-only QA Prompt for Qwen2.5-7B-Instruct and Llama-3.1-8B-Instruct,
  fonttitle=\bfseries,
  breakable,
  width=\textwidth,
  left=3pt,
  right=3pt,
  top=3pt,
  bottom=3pt,
  boxsep=2pt,
]
{\footnotesize
\begin{lstlisting}[breaklines=true, columns=flexible]
You are a question answering model, your task is to answer the question based on factual information in your own knowledge.

Please answer with a single entity, as concise as possible.

Please reason step by step and then provide the final answer. The reasoning process must be enclosed within <think> </think> tags. The final answer MUST be put in \boxed{}. For example, \boxed{3 times}, \boxed{New York}, etc.

Question: {question}
Answer:
\end{lstlisting}
}
\end{tcolorbox}

TruthRL and UCPO use an abstention-enabled version to elicit the model's internal abstention behavior:
\begin{tcolorbox}[
  colback=gray!5,
  colframe=gray!75!black,
  title=Abstention-enabled QA Prompt for Qwen2.5-7B-Instruct and Llama-3.1-8B-Instruct,
  fonttitle=\bfseries,
  breakable,
  width=\textwidth,
  left=3pt,
  right=3pt,
  top=3pt,
  bottom=3pt,
  boxsep=2pt,
]
{\footnotesize
\begin{lstlisting}[breaklines=true, columns=flexible]
You are a question answering model, your task is to answer the question based on factual information in your own knowledge.

Please adhere to the following guidelines when formulating the answer:
1. Answer with a single entity, as concise as possible.
2. If you are uncertain or don't know the answer, answer "I don't know".

Please reason step by step and then provide the final answer. The reasoning process must be enclosed within <think> </think> tags. The final answer MUST be put in \boxed{}. For example, \boxed{I don't know}, \boxed{3 times}, \boxed{New York}, etc.

Question: {question}
Answer:
\end{lstlisting}
}
\end{tcolorbox}

\paragraph{Prompts for Qwen3-8B.}
\label{app:prompt-qwen3}
Since thinking tags are provided in the chat template of Qwen3-8B and prompting it to generate these tags can violate its generation format, we omit explicit requirements to generate thinking tags.

\begin{tcolorbox}[
  colback=gray!5,
  colframe=gray!75!black,
  title=Answer-only QA Prompt for Qwen3-8B,
  fonttitle=\bfseries,
  breakable,
  width=\textwidth,
  left=3pt,
  right=3pt,
  top=3pt,
  bottom=3pt,
  boxsep=2pt,
]
{\footnotesize
\begin{lstlisting}[breaklines=true, columns=flexible]
You are a question answering model, your task is to answer the question based on factual information in your own knowledge.

Please answer with a single entity, as concise as possible.

Think carefully before answering. Output only the final answer in \boxed{}. For example, \boxed{3 times}, \boxed{New York}, etc.

Question: {question}
Answer:
\end{lstlisting}
}
\end{tcolorbox}

Its abstention-enabled counterpart is:

\begin{tcolorbox}[
  colback=gray!5,
  colframe=gray!75!black,
  title=Abstention-enabled QA Prompt for Qwen3-8B,
  fonttitle=\bfseries,
  breakable,
  width=\textwidth,
  left=3pt,
  right=3pt,
  top=3pt,
  bottom=3pt,
  boxsep=2pt,
]
{\footnotesize
\begin{lstlisting}[breaklines=true, columns=flexible]
You are a question answering model, your task is to answer the question based on factual information in your own knowledge.

Please adhere to the following guidelines when formulating the answer:
1. Answer with a single entity, as concise as possible.
2. If you are uncertain or don't know the answer, answer "I don't know".

Think carefully before answering. Output only the final answer in \boxed{}. For example, \boxed{I don't know}, \boxed{3 times}, \boxed{New York}, etc.

Question: {question}
Answer:
\end{lstlisting}
}
\end{tcolorbox}

\paragraph{Correctness and abstention judge.}
\label{app:prompt-judge}
The following is the complete user message sent to the judge model.
It is shared by reward computation and evaluation.

\begin{tcolorbox}[
  colback=gray!5,
  colframe=gray!75!black,
  title=Prompt for LLM-as-a-Judge,
  fonttitle=\bfseries,
  breakable,
  width=\textwidth,
  left=3pt,
  right=3pt,
  top=3pt,
  bottom=3pt,
  boxsep=2pt,
]
{\footnotesize
\begin{lstlisting}[breaklines=true, columns=flexible]
Your job is to look at a question, a gold target, and a predicted answer, and then assign a grade of either ["CORRECT", "INCORRECT", "ABSTENTION"].
The predicted answer you receive is the final extracted answer only, not the model's chain-of-thought.
Non-negotiable rule: if the predicted answer is a placeholder, template slot, anonymized variable, or missing-value marker instead of the actual requested answer, grade it as INCORRECT. It does not matter whether the placeholder has the right answer type.
First, I will give examples of each grade, and then you will grade a new example.

The following are examples of CORRECT predicted answers.
```
Question: What are the names of Barack Obama's children?
Gold target: Malia Obama and Sasha Obama
Predicted answer 1: sasha and malia obama
Predicted answer 2: Barack Obama has two daughters. Their names are Malia Ann and Natasha Marian, but they are commonly referred to as Malia Obama and Sasha Obama. Malia was born on July 4, 1998, and Sasha was born on June 10, 2001.
```
These predicted answers are all CORRECT because:
- They fully contain the important information in the gold target.
- They do not contain any information that contradicts the gold target.
- Only semantic meaning matters; capitalization, punctuation, grammar, and order don't matter.
- Hedging and guessing are not allowed.

The following are examples of INCORRECT predicted answers.
```
Question: What are the names of Barack Obama's children?
Gold target: Malia and Sasha
Predicted answer 1: Malia.
Predicted answer 2: Malia, Sasha, and Susan.
Predicted answer 3: Barack Obama does not have any children.
Predicted answer 4: I think it's either Malia and Sasha. Or it could be Malia and Jackie. Or it could be Joey and Malia.
Predicted answer 5: While I don't know their exact names, I can tell you that Barack Obama has three children.
Predicted answer 6: <NAME>
Predicted answer 7: [person]
Predicted answer 8: <the answer>
```
These predicted answers are all INCORRECT because:
- A factual statement in the answer contradicts the gold target.
- The response is not an answer, or is an answer to another question.
- Incorrect statements that have some hedging (e.g., "it is possible that", "although i'm not sure, i think") are also considered incorrect.
- A placeholder, template slot, anonymized variable, or missing-value marker is not an answer, even if it has the right type.

The following are examples of ABSTENTION predicted answers.
```
Question: What are the names of Barack Obama's children?
Gold target: Malia and Sasha
Predicted answer 1: I don't know.
Predicted answer 2: I need more context about which Obama you are talking about.
Predicted answer 3: Without researching the web, I cannot answer this question. However, I can tell you that Barack Obama has two children.
Predicted answer 4: Barack Obama has two children. I know that one of them is Malia, but I'm not sure about the other one.
```
These predicted answers are all ABSTENTION because:
- They explicitly state inability to answer, and no statements in the answer contradict the gold target.
- They explicitly state that the question is unanswerable or fictional.

Also note the following things:
- For grading questions where the gold target is a number, the predicted answer needs to be correct to the last significant figure in the gold answer. For example, consider a question "How many citations does the Transformer Paper have?" with gold target "120k".
- Predicted answers "120k", "124k", and "115k" are all CORRECT.
- Predicted answers "100k" and "113k" are INCORRECT.
- The gold target may contain more information than the question. In such cases, the predicted answer only needs to contain the information that is in the question.
- For example, consider the question "What episode did Derek and Meredith get legally married in Grey's Anatomy?" with gold target "Season 7, Episode 20: White Wedding". Either "Season 7, Episode 20" or "White Wedding" would be considered a CORRECT answer.
- Do not punish predicted answers if they omit information that would be clearly inferred from the question.
- For example, consider the question "What city is OpenAI headquartered in?" and the gold target "San Francisco, California". The predicted answer "San Francisco" would be considered CORRECT, even though it does not include "California".
- Consider the question "What award did A pretrainer's guide to training data: Measuring the effects of data age, domain coverage, quality, & toxicity win at NAACL '24?", the gold target is "Outstanding Paper Award". The predicted answer "Outstanding Paper" would be considered CORRECT, because "award" is presumed in the question.
- For the question "What is the height of Jason Wei in meters?", the gold target is "1.73 m". The predicted answer "1.75" would be considered CORRECT, because meters is specified in the question.
- For the question "What is the name of Barack Obama's wife?", the gold target is "Michelle Obama". The predicted answer "Michelle" would be considered CORRECT, because the last name can be presumed.
- Do not punish for typos in people's name if it's clearly the same name.
- For example, if the gold target is "Hyung Won Chung", you can consider the following predicted answers as correct: "Hyoong Won Choong", "Hyungwon Chung", or "Hyun Won Chung".
- Always grade placeholders as INCORRECT, not CORRECT or ABSTENTION. A placeholder is text that stands in for a missing answer instead of providing the actual entity, value, title, place, genre, date, or number.

Grade the predicted answer of this new question as one of:
A: CORRECT
B: INCORRECT
C: ABSTENTION
Respond with exactly one of the letters "A", "B", or "C", with no punctuation, asterisks or text before or after it.

Here is a new example. Simply reply with either "A", "B", "C". Don't apologize or correct yourself if there was a mistake; we are just trying to grade the answer.
Before giving the verdict, apply this check: if the predicted answer is a placeholder or template slot rather than a concrete answer, the verdict must be B.
```
Question: {question}
Gold target: {target}
Predicted answer: {predicted_answer}
```
Verdict:
\end{lstlisting}
}
\end{tcolorbox}

%% file: tables/Appendix_data.tex
\begin{wraptable}{r}{0.4\textwidth}
    \centering
    \caption{Dataset partitions.}
    \label{tab:setup-data-splits}
    \begin{adjustbox}{max width=0.4\textwidth}
    \begin{tabular}{lrrr}
        \toprule
        \textbf{Dataset} & \textbf{Training} & \textbf{Validation} & \textbf{Test} \\
        \midrule
        NQ       & 10,000 & 512 & 3,610 \\
        PopQA    & 5,000  & 256 & 9,011 \\
        TriviaQA & 10,000 & 512 & 11,313 \\
        \bottomrule
    \end{tabular}
    \end{adjustbox}
\end{wraptable}

%% file: tables/Appendix_training.tex
\begin{table}[t]
    \centering
    \small
    \setlength{\tabcolsep}{4pt}
    \caption{RL training hyperparameters.}
    \label{tab:setup-rl-hyperparameters}
    \begin{tabularx}{\linewidth}{@{}>{\raggedright\arraybackslash}p{0.33\linewidth}
                                  >{\raggedright\arraybackslash}X
                                  >{\raggedright\arraybackslash}X@{}}
        \toprule
        \multirow{2}{*}{\textbf{Setting}} &
        \textbf{Qwen2.5-7B-Instruct / \newline Llama-3.1-8B-Instruct} &
        \multirow{2}{*}{\textbf{Qwen3-8B}} \\
        \midrule
        Training budget
            & 8 epochs & 8 epochs \\
        Questions per rollout batch
            & 128 & 128 \\
        Responses per question
            & 5 & 5 \\
        Rollout temperature / top-$p$
            & 1.0 / 1.0 & 1.0 / 1.0 \\
        Rollout top-$k$ truncation
            & Disabled & Disabled \\
        Maximum prompt length
            & 512 tokens & 512 tokens \\
        Native thinking
            & Not applicable & Enabled \\
        Maximum response length
            & 256 tokens & 2048 tokens \\
        Optimizer
            & AdamW & AdamW \\
        Learning rate
            & $10^{-6}$ & $10^{-6}$ \\
        Schedule / warmup
            & Constant / none & Constant / none \\
        Weight decay
            & 0.01 & 0.01 \\
        AdamW $(\beta_1,\beta_2)$
            & $(0.9,0.999)$ & $(0.9,0.999)$ \\
        PPO mini-batch size
            & 64 prompts & 64 prompts \\
        Micro-batch size
            & 16 trajectories per GPU & 2 trajectories per GPU \\
        PPO epochs per rollout batch
            & 1 & 1 \\
        Clipping ratio
            & 0.2 & 0.2 \\
        Gradient-norm clipping
            & 1.0 & 1.0 \\
        Loss aggregation
            & Token mean & Token mean \\
        KL coefficient
            & 0.001 & 0.001 \\
        Entropy-bonus coefficient
            & 0 & 0 \\
        Checkpoint / validation interval
            & Every 100 steps & Every 50 steps \\
        Checkpoint used
            & The last checkpoint & The last checkpoint \\
        \bottomrule
    \end{tabularx}
\end{table}

%% file: tables/Appendix_judge.tex
\begin{table*}[t]
\centering
\caption{
Detailed reliability analysis of LLM judges against human annotations ($n=100$ per stage).
TruthRL uses zero abstention reward.
Agreement: the percentage of exact label matches;
False acceptance: human $\times$/$\bot$ labeled $\checkmark$ by the judge;
False rejection: human $\checkmark$ labeled $\times$/$\bot$.
$\times$/$\bot$ confusion: swaps between incorrect and abstention.}
\begin{adjustbox}{max width=\textwidth}
\begin{tabular}{llrrrrr}
\toprule
& & \textbf{gpt-oss-20b} & \textbf{DeepSeek-V4.1-Flash} & \textbf{GPT 5.5} & \textbf{Gemini 3.6 Flash} & \textbf{Claude Sonnet 5} \\
\midrule
\multirow{4}{*}{\rotatebox[origin=c]{90}{\textbf{PreRL}}}
& Agreement (\%) & 91.0 & 92.0 & 95.0 & 92.0 & 97.0 \\
& False acceptance & \textbf{0} & 3 & 3 & 7 & 2 \\
& False rejection & 8 & 3 & 2 & 0 & 1 \\
& $\times$/$\bot$ confusion & 1 & 2 & 0 & 1 & 0 \\
\midrule
\multirow{4}{*}{\rotatebox[origin=c]{90}{\textbf{AccRL}}}
& Agreement (\%) & 90.0 & 94.0 & 91.0 & 91.0 & 96.0 \\
& False acceptance & \textbf{1} & 5 & 2 & 8 & 2 \\
& False rejection & 9 & 1 & 7 & 1 & 2 \\
& $\times$/$\bot$ confusion & 0 & 0 & 0 & 0 & 0 \\
\midrule
\multirow{4}{*}{\rotatebox[origin=c]{90}{\textbf{TruthRL}}}
& Agreement (\%) & 92.0 & 97.0 & 92.0 & 94.0 & 93.0 \\
& False acceptance & \textbf{0} & 3 & 3 & 6 & 4 \\
& False rejection & 8 & 0 & 5 & 0 & 3 \\
& $\times$/$\bot$ confusion & 0 & 0 & 0 & 0 & 0 \\
\bottomrule
\end{tabular}
\end{adjustbox}
\label{tab:judge-reliability}
\end{table*}

%% file: tables/Appendix_decomposition.tex
\begin{table*}[t]
    \centering
    \small
    \caption{Answering and distinguishing ability at matched coverage for
    every model--dataset setting. All entries are percentages.
    Selective answering matches TruthRL's coverage within each setting.
    The highest value for each metric within each row is in \textbf{bold}.}
    \label{tab:app-decomposition}
    \setlength{\tabcolsep}{3pt}
    \begin{tabular}{@{}llr rr rr rr rr@{}}
        \toprule
        & & & \multicolumn{2}{c}{\textbf{TruthRL}} & \multicolumn{6}{c}{\textbf{Selective Answering}} \\
        \cmidrule(lr){4-5}\cmidrule(lr){6-11}
        \textbf{Model} & \textbf{Dataset} & \textbf{Cov.} & & & \multicolumn{2}{c}{Vanilla}
        & \multicolumn{2}{c}{P(True)} & \multicolumn{2}{c}{S.E.} \\
        \cmidrule(lr){6-7}\cmidrule(lr){8-9}\cmidrule(lr){10-11}
        & & & $G_{\mathrm{ans}}$ & $G_{\mathrm{dis}}$
        & $G_{\mathrm{ans}}$ & $G_{\mathrm{dis}}$
        & $G_{\mathrm{ans}}$ & $G_{\mathrm{dis}}$
        & $G_{\mathrm{ans}}$ & $G_{\mathrm{dis}}$ \\
        \midrule
        \multirow{3}{*}{Qwen2.5-7B-Instruct} & NQ & 50.60 & \textbf{6.46} & 87.36 & 0.00 & 67.93 & 0.00 & 82.63 & 0.00 & \textbf{88.07} \\
         & PopQA & 19.69 & \textbf{5.30} & 91.75 & 0.00 & 78.91 & 0.00 & 88.43 & 0.00 & \textbf{92.03} \\
         & TriviaQA & 72.70 & \textbf{6.13} & 86.23 & 0.00 & 43.75 & 0.00 & 75.42 & 0.00 & \textbf{86.83} \\
        \midrule
        \multirow{3}{*}{Qwen3-8B} & NQ & 50.45 & \textbf{4.74} & 84.89 & 0.00 & 65.34 & 0.00 & 78.96 & 0.25 & \textbf{85.59} \\
         & PopQA & 20.07 & \textbf{1.19} & 88.96 & 0.00 & 76.01 & 0.00 & 86.45 & 0.00 & \textbf{90.00} \\
         & TriviaQA & 78.13 & \textbf{3.29} & \textbf{83.66} & 0.00 & 39.69 & 0.32 & 65.96 & 0.00 & 83.57 \\
        \midrule
        \multirow{3}{*}{Llama-3.1-8B-Instruct} & NQ & 77.48 & \textbf{6.61} & \textbf{78.17} & 0.00 & 59.39 & 0.00 & 68.21 & 0.00 & 77.25 \\
         & PopQA & 43.37 & \textbf{9.12} & \textbf{86.89} & 0.00 & 69.69 & 0.00 & 80.91 & 0.00 & 86.37 \\
         & TriviaQA & 90.10 & \textbf{2.78} & 70.67 & 0.00 & 34.49 & 0.00 & 46.61 & 0.00 & \textbf{72.02} \\
        \midrule
        \multicolumn{2}{l}{Avg.} & 55.84 & \textbf{5.07} & 84.29 & 0.00 & 59.47 & 0.04 & 74.84 & 0.03 & \textbf{84.64} \\
        \bottomrule
    \end{tabular}
\end{table*}